\documentclass[twocolumn, switch]{article}

\usepackage{preprint}

\usepackage[utf8]{inputenc}
\usepackage[T1]{fontenc}

\usepackage{amsmath, amsthm, amssymb, amsfonts}

\usepackage[numbers,square]{natbib}
\usepackage{xcolor}
\usepackage{nicefrac}
\usepackage{microtype}
\usepackage{lineno}
\usepackage{float}
\usepackage{lipsum}
\usepackage{tikz}
\usepackage{todonotes}
\usepackage{adjustbox}
\usepackage{graphicx}
\usepackage{stfloats}
\graphicspath{{./images/}}
\usepackage{graphicx}
\usepackage{subcaption}
\usepackage{adjustbox}
\usepackage{booktabs}
\usepackage{multirow}
\usepackage{tabularx}

\usepackage[acronym]{glossaries}

\newacronym{sits}{SITS}{Satellite Image Time Series}
\newacronym{rf}{RF}{Random Forest}
\newacronym{xgb}{XGBoost}{Extreme Gradient Boosting}
\newacronym{lstm}{LSTM}{Long Short-Term Memory}
\newacronym{gru}{GRU}{Gated Recurrent Unit}
\newacronym{ig}{IG}{Integrated Gradients}
\newacronym{mha}{MHA}{Multi-head Attention}
\newacronym{rnn}{RNN}{Recurrent Neural Network}
\newacronym{doy}{DOY}{day of year}

\usepackage{newfloat}
\DeclareFloatingEnvironment[name={Supplementary Figure}]{suppfigure}

\usepackage{sidecap}
\sidecaptionvpos{figure}{c}

\usepackage{titlesec}
\titlespacing\section
  {0pt}{12pt plus 3pt minus 3pt}{1pt plus 1pt minus 1pt}
\titlespacing\subsection
  {0pt}{10pt plus 3pt minus 3pt}{1pt plus 1pt minus 1pt}
\titlespacing\subsubsection
  {0pt}{8pt plus 3pt minus 3pt}{1pt plus 1pt minus 1pt}

\usepackage[
  colorlinks=true,
  linkcolor=purple,
  urlcolor=blue,
  citecolor=cyan,
  anchorcolor=black
]{hyperref}

\definecolor{lime}{HTML}{A6CE39}
\DeclareRobustCommand{\orcidicon}{
	\begin{tikzpicture}
	\draw[lime, fill=lime] (0,0)
	circle [radius=0.16]
	node[white] {{\fontfamily{qag}\selectfont \tiny ID}};
	\draw[white, fill=white] (-0.0625,0.095)
	circle [radius=0.007];
	\end{tikzpicture}
	\hspace{-2mm}
}
\foreach \x in {A, ..., Z}{\expandafter\xdef\csname orcid\x\endcsname{\noexpand\href{https://orcid.org/\csname orcidauthor\x\endcsname}
			{\noexpand\orcidicon}}
}

\title{Exploring Interpretability in Deep Learning-Based Forest Classification - Separating Similar Forest Classes using Sentinel-2 Time Series}

\usepackage{xwatermark}
\newwatermark[firstpage,color=gray!60,angle=90,scale=0.32, xpos=-4.05in,ypos=0]{\href{https://doi.org/}{\color{gray}{Publication doi}}}
\newwatermark[firstpage,color=gray!60,angle=90,scale=0.32, xpos=3.9in,ypos=0]{\href{https://doi.org/}{\color{gray}{Preprint doi}}}
\newwatermark[firstpage,color=gray!90,angle=0,scale=0.28, xpos=0in,ypos=-5in]{*correspondence: \texttt{maximilian.kirsch@h-ka.de}}

\usepackage{authblk}

\author[1]{Maximilian Kirsch\orcidA{}\thanks{\texttt{maximilian.kirsch@h-ka.de}}}
\author[1]{Christine Preisach\orcidB{}}

\affil[1]{%
    Intelligent Systems Research Group (ISRG),
    Karlsruhe University of Applied Sciences, Germany
}

\begin{document}

\twocolumn[ 
  \begin{@twocolumnfalse} 

\maketitle

\begin{abstract}
The ecological insurance theory posits that diverse forests are more resilient and better able to adapt to changing environmental conditions. Increasing drought and heat stress in Central Europe promotes mixed forests for maintaining ecosystem services.
However, a diverse species distribution complicates remote sensing-based species mapping, as distinguishing pure from mixed species compositions at the pixel level remains challenging. We evaluated the potential of recurrent neural networks (RNNs) for tree-species classification from Sentinel-2 time series and investigated the spectral-temporal features underlying class discrimination, with particular emphasis on separating pure from mixed Norway spruce pixels.
We compared RNN classification models with Random Forest (RF) and Extreme Gradient Boosting (XGBoost) and used Integrated Gradients (IG) to quantify feature attributions across spectral bands and time. The best-performing RNN achieved a Macro-F1 score of 0.851 on the test data, outperforming XGBoost by 0.012 and Random Forest by 0.103. Beyond classification performance, the attribution analysis revealed distinct spectral-temporal patterns associated with the separation of pure and mixed Norway spruce pixels. Attribution values increased particularly after approximately day of year (DOY) 125, while B11 showed the strongest contribution during periods of increasing spectral divergence between the two classes.
These findings show that recurrent models can provide competitive classification performance while also revealing the phenological periods and spectral regions that are most relevant for distinguishing closely related forest compositions. Combining time-series classification with explainability therefore provides additional insight into the spectral-temporal mechanisms underlying tree-species discrimination and may support more detailed mapping of mixed forest structures.
\end{abstract}

\vspace{0.35cm}

  \end{@twocolumnfalse} 
] 

\section{Introduction}
Understanding the spatial distribution of tree species provides a basis for decision-making in forest management, biodiversity conservation, and climate protection \citep{zhong_review_2024}. Repeated monitoring of forest composition can further reveal changes caused by natural disturbances, harvesting, and regrowth, thereby supporting forest inventories and long-term forest management \citep{stoffels_satellite-based_2012, tan_leveraging_2025}. Under increasingly frequent heat, drought, and disturbance events \citep{seneviratne_weather_2021}, forests in Central Europe need to be more resilient and adapted to change \citep{pretzsch_mixed-species_2017}. The ecological insurance theory posits that greater species diversity can stabilize ecosystem functioning to environmental fluctuations \citep{yachi_biodiversity_1999}. Hence, the \textit{New EU Forest Strategy for 2030} of the European Commission explicitly promotes the creation and maintenance of functionally diverse, mixed-species forests \citep{european_commission_joint_research_centre_climate_2020}. Consequently, mapping approaches must also reliably represent mixed species compositions. 
Satellite-based forest classification has been applied across a wide range of spatial scales, from site-level studies to global forest mapping \citep{iglseder_potential_2023, martone_global_2018}. Sentinel-2 is particularly well suited for this purpose because it provides freely available multispectral imagery at high revisit frequency and spatial resolutions of up to 10 m. Early applications largely relied on single-date imagery. For example, \citet{immitzer_first_2016} evaluated Sentinel-2 data for crop and tree species classification in Central Europe using a supervised \gls*{rf} classifier and achieved an overall accuracy of 65\% for tree species classification. Their results further highlighted the importance of the red-edge and short-wave infrared (SWIR) bands. However, single-date observations capture only one stage of the seasonal development and may therefore provide limited discriminatory information for species with similar spectral characteristics at a given phenological stage. With the launch of Sentinel-2B in 2017, the increased observation frequency facilitated the use of \gls{sits} for vegetation and forest classification \citep{phiri_sentinel-2_2020}. Exploiting this temporal information can improve species discrimination by capturing phenological differences in seasonal spectral trajectories. \citet{grabska_forest_2019}, for example, compared single-date and multi-temporal Sentinel-2 imagery for the classification of nine tree species in mixed forests and reported an increase in overall mapping accuracy of 5–10\% when using time-series data, with combinations of spring and autumn observations being particularly informative.  
In tree species classification, Sentinel-2 \gls{sits} are commonly analyzed using machine-learning approaches, with tree-based ensemble methods such as \gls*{rf} and \gls*{xgb} being frequently applied (e.g., \cite{grabska_forest_2019, blickensdorfer_national_2024, wegler_tree_2025, liu_evaluating_2024, abdi_mapping_2026}). These methods are widely used because of their strong predictive performance, scalability to large study areas, and comparatively high interpretability \citep{mouret_tree_2025}. However, SITS contain sequential spectral information whose temporal dependencies can also be modeled directly using deep learning approaches. The continued operation of Sentinel-2 has progressively expanded the available image archive, providing increasingly long time series that also facilitate the application of such sequence-based deep-learning approaches (e.g.,\cite{freudenberg_consistent_2025, mouret_tree_2025}). 

Despite their increasing relevance for ecology and management, distinguishing mixed forests from pure stands using remote sensing data is challenging \citep{abdullah_timing_2019}. In medium-resolution satellite images, such as  Sentinel-2 (10m), pixels are likely to contain several individual trees, and the proportion of each species within the pixels can vary substantially, which is reflected in classification performance. \citet{blickensdorfer_national_2024}, for example, mapped the main tree species in Germany using combined Sentinel-1 and Sentinel-2 time series and integrated exogenous features such as temperature and topographic features. While they reported high F1 scores for the five dominant species in pure stands, ranging from 72\% to 97\%, the inclusion of mixed stands reduced accuracy by 4-14 percentage points. At the same time, most forestry inventory, planning, and management tools have been developed for monocultures \citep{pretzsch_mixed-species_2017} and tree species classification studies prominently focus on pure stands (e.g., \citep{wegler_tree_2025, freudenberg_sentinel-2_2025}). This is a limitation, especially in Germany, as mixed forests account for 79\% of Germany’s forest area in 2022 \citep{bundesministerium_fur_landwirtschaft_ernahrung_und_heimat_bundeswaldinventur_2022}. Disregarding mixed stands in forest mapping products can therefore substantially misrepresent forest composition. 
Given the difficulty of distinguishing mixed from pure forest classes, classification performance alone provides limited insight into which spectral and temporal characteristics enable their separation. Research on explainability in remote sensing-based tree species classification has predominantly focused on tree-based methods, partly for their comparatively high interpretability (e.g., \cite{wegler_tree_2025, immitzer_optimal_2019, liu_evaluating_2024}; \citep{mouret_tree_2025}). \citet{immitzer_optimal_2019} investigated the spectral variables and acquisition dates most relevant for classifying individual tree species using \gls{rf} models trained on combinations of Sentinel-2 bands and spectral indices across 18 scenes. Their results highlighted the red and short-wave infrared bands, while observations acquired between April and August contributed most strongly to classification performance. Similarly, \citet{wegler_tree_2025} assessed feature contributions in \gls{rf} and \gls{xgb} models using SHAP values and model-based feature importance and identified the short-wave infrared and red-edge bands among the most influential features. While \citet{malkanthi_forest_2025} also addressed explainability in deep learning-based tree species classification, their analysis focused on exogenous variables rather than on the spectral-temporal information contained in the SITS.
Consequently, although previous studies have provided valuable insights into the spectral and temporal features relevant for tree species classification, the spectral-temporal decision processes of deep learning models remain insufficiently understood. This limitation is particularly relevant for closely related classes, such as pure and mixed Norway spruce, for which understanding the wavelengths and phenological periods driving class separation may provide insights beyond overall classification performance. \\

To address this gap in \gls{sits}-based tree species classification, we propose an explainability workflow for multiclass classification that identifies the spectral-temporal features most relevant for class discrimination while remaining applicable to deep learning models. We particularly investigate the separation of pure and mixed Norway spruce pixels and additionally examine deciduous tree species with similar phenological patterns. The study addresses the following research questions: 

\textbf{RQ1:} How do recurrent deep learning models compare with established tree-based ensemble models in classifying tree species and distinguishing pure from mixed forest compositions using one-year Sentinel-2 time series? 

\textbf{RQ2:} Can attribution techniques applied to deep learning models provide insight into the spectral bands and phenological periods relevant for class discrimination, particularly between closely related forest classes?

\section{Materials and Methods}
\subsection{Dataset and Study Area}

\begin{figure*}
\centering
\includegraphics[width=0.8\textwidth]{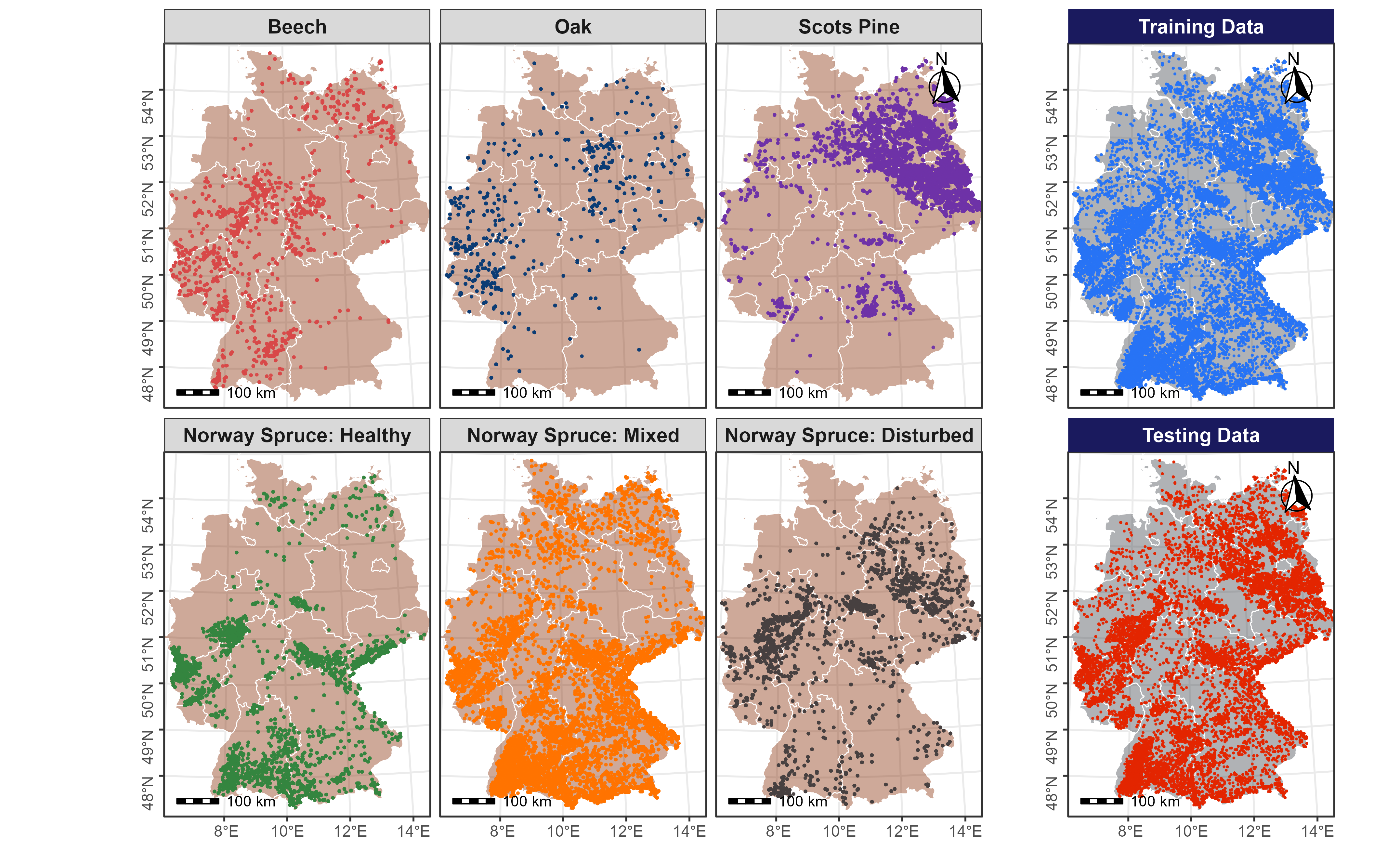}
    \caption{\textit{Data distribution}: spatial distribution of sample locations used in this study, grouped by species.}
    \label{fig:placeholder}
\end{figure*}

We used the publicly available Sentinel-2 tree species time-series dataset for Germany, published in 2025 by \citet{freudenberg_sentinel-2_2025}. The dataset combines bottom-of-atmosphere reflectance time series from analysis-ready Sentinel-2 imagery with tree species labels derived from the German National Forest Inventory. It covers individual tree observations across Germany from 2015 to October 2022 and includes 48 tree species and three species groups. 
We selected the four dominant tree species or species groups in Germany: Norway spruce (Picea abies), Scots pine (\textit{Pinus sylvestris}), European beech (\textit{Fagus sylvatica}), and oak, with the latter comprising sessile oak (\textit{Quercus petraea}) and English oak (\textit{Quercus robur}) \citep{thunen-institut_bundeswaldinventur_2024}. Furthermore, we split the Norway spruce class into healthy and disturbed Norway spruce, as well as mixed pixels that comprise Norway spruce and other tree species (mainly beech) in mixed stands. Additional classes were deliberately excluded, as the remaining tree species in the dataset by \citet{freudenberg_sentinel-2_2025} were underrepresented for the study's specific use case of classification and separability analysis.
An overview of the data basis and the spatial distribution of the dataset locations is shown in Figure~\ref{fig:placeholder}.

\begin{figure}
\centering
\includegraphics[width=1\columnwidth]{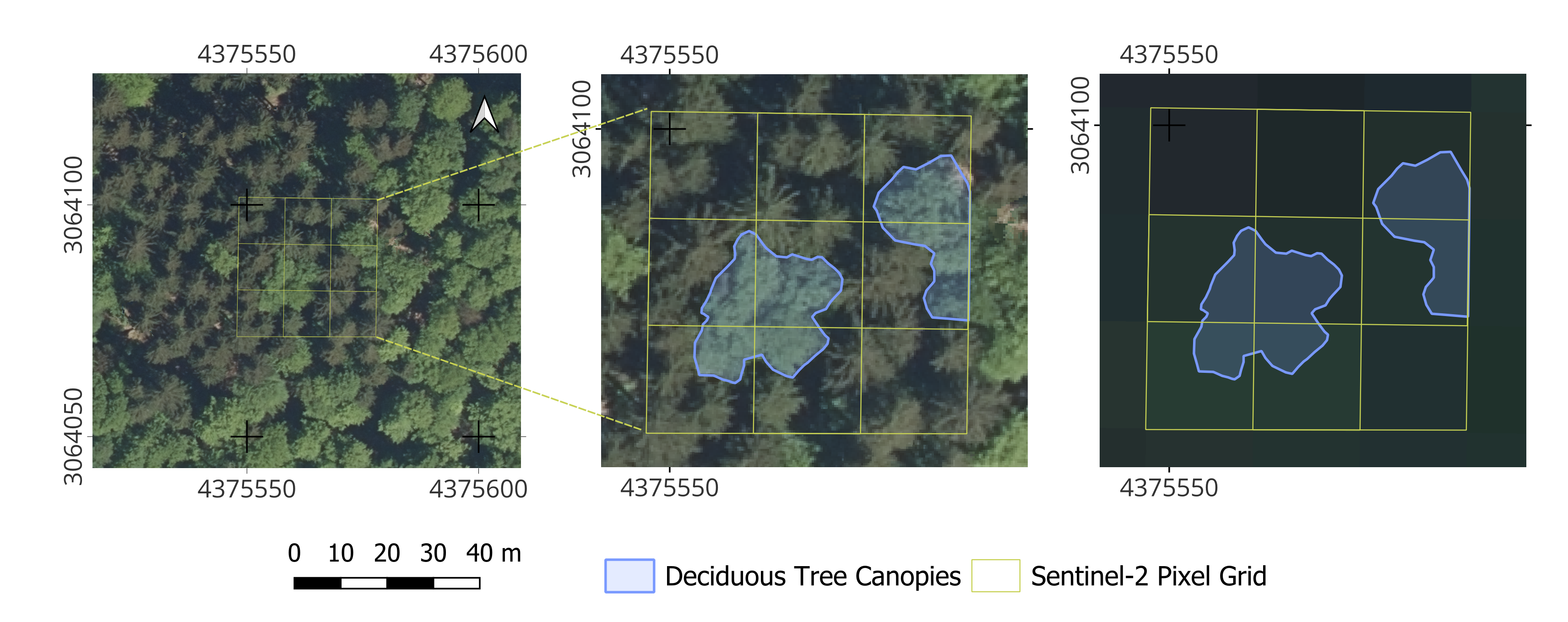}
    \caption{\textit{Mixed Sentinel-2 pixels.} Illustration of Sentinel-2 pixels containing pure Norway spruce stands and mixed stands. \textit{A} shows an orthophoto of a forest structure including Norway spruce and varying deciduous trees. \textit{B} shows a subset of the same area including the outlines of deciduous trees within a 3~$\times$~3 Sentinel-2 pixel grid. \textit{C} shows the corresponding Sentinel-2 pixels averaged over one year (2021).}
    \label{fig:mixpix}
\end{figure}

The selected dataset comprised pure and mixed Norway spruce time series, as our objective was to determine whether the models could differentiate between classes that are very closely related. Most tree-species maps in Germany focus on pure stands because mixed pixels introduce substantial classification uncertainty. Their spectral signatures tend to resemble those of the dominant tree species within the pixel, which can lead to mixed pixels being misclassified as pure stands (see Figure~\ref{fig:mixpix}).

We subsequently filtered the selected data and applied the preprocessing steps described in Section ~\ref{chap31}.

\newpage
\subsection{Models}

\begin{figure}
    \centering
    \includegraphics[width=0.65\columnwidth]{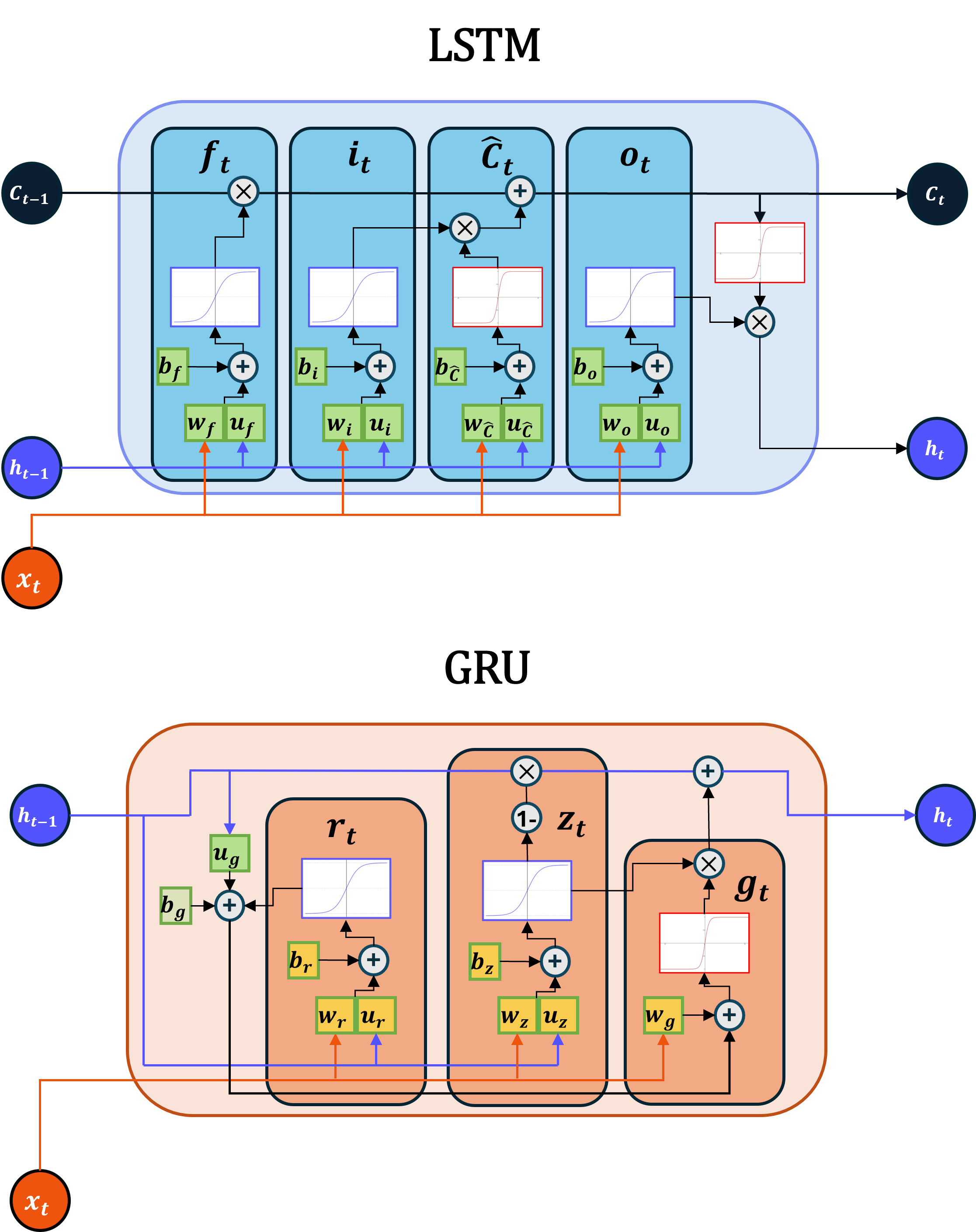}

\caption{\textit{LSTM and GRU architectures.} At time step \(t\), \(x_t\) is the multivariate input. The LSTM unit uses forget (\(f_t\)), input (\(i_t\)), and output (\(o_t\)) gates with candidate cell state \(\tilde{C}_t\). The GRU unit uses update (\(z_t\)) and reset (\(r_t\)) gates with candidate activation \(g_t\). The LSTM carries forward long- and short-term memory separately as cell state \(C\) and hidden state \(h\), while GRU only forwards one combined state.  Blue and red frames indicate sigmoid (\(\sigma\)) and \(\tanh\) activations, respectively. The recurrent weights \(w,u\) and biases \(b\) are learned.}
    \label{fig:lstmGRU}
\end{figure}

In this study, we selected two ensemble methods, \gls*{xgb} and \gls*{rf}, as well as four deep learning approaches. 
\gls*{xgb} and \gls*{rf} were selected because they have proven to be very successful in the field of SITS forest classification \citep{wegler_tree_2025, liu_evaluating_2024, blickensdorfer_national_2024}.  
For the deep learning models, we developed an architecture that utilizes \glspl{rnn}, which are particularly well suited to sequential data because they can model temporal dependencies within multivariate time series \citep{russell_artificial_2020}. Their potential for \gls{sits} data has also been explored and validated (e.g., \citep{kirsch_early_2025, kong_long_2018}). We selected \gls{lstm} \citep{hochreiter_long_1997} and \gls{gru} \citep{cho_learning_2014} to extract spectral-temporal information from Sentinel-2 time series and embed it into a latent feature space. Their schematic functionality is illustrated in Figure~\ref{fig:lstmGRU}. 

The equations for the gated units and the forward pass of the \gls{lstm} and \gls{gru} are provided in Appendix~\ref{app:lstmandgru_equations}. 

\begin{figure*}
    \centering
    \includegraphics[width=0.8\textwidth]{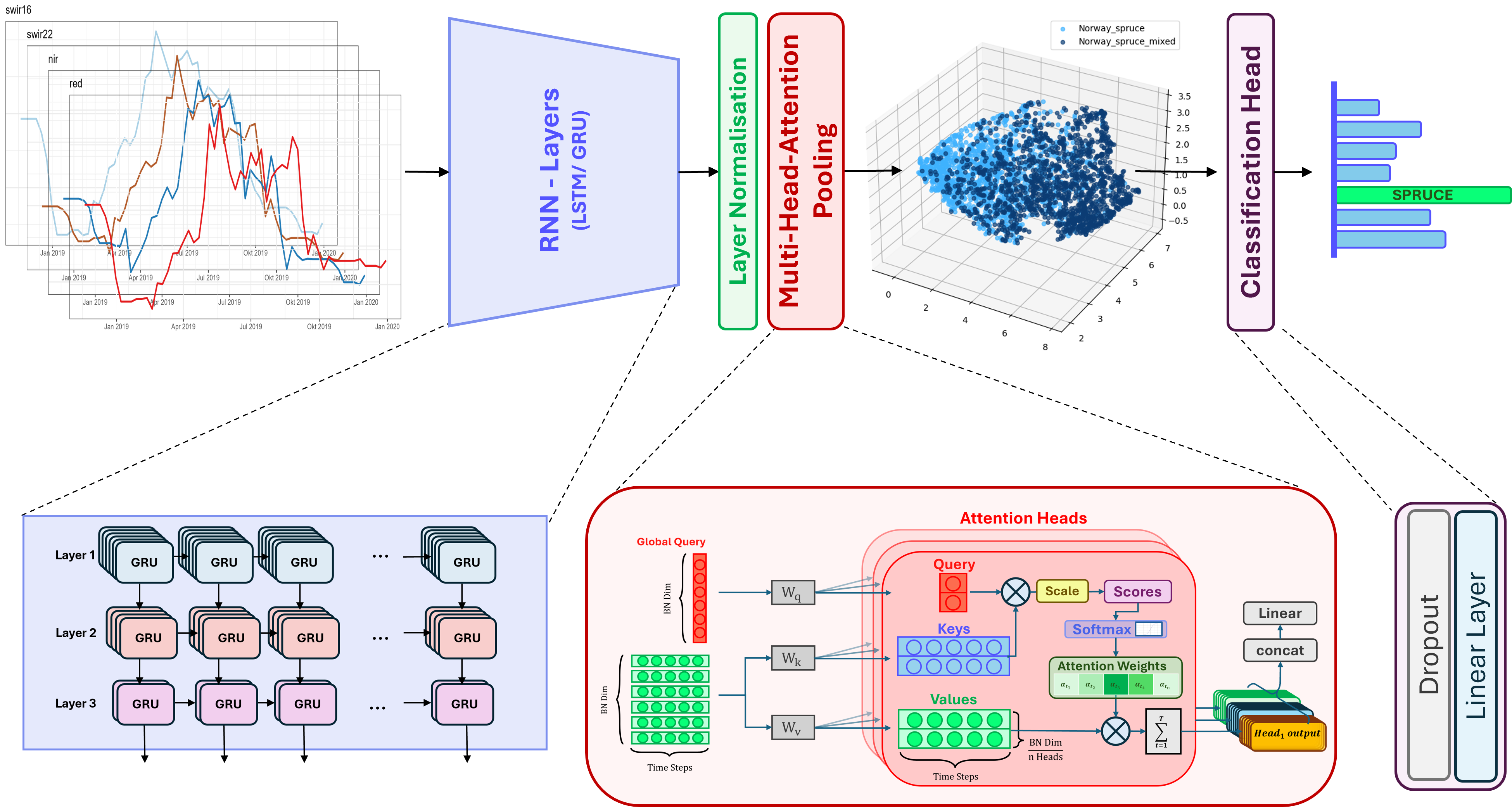}
    \caption{\textit{GRU-MHA Classifier Architecture.} The RNN output is projected into key and value representations, while a learnable global query provides the query representation. After splitting the projected representations into multiple attention heads, scaled dot-product attention is computed between the global query and the keys of all time steps. The resulting attention scores are normalized using softmax and used to calculate a weighted sum of the value vectors, yielding one pooled representation per head. These are connected and used as inputs for the classification head.}
    \label{fig:lstmmhaarchitectureLS}
\end{figure*}

The general model architecture follows an RNN-encoder design, specifically the encoder described by \citet{kirsch_early_2025}. The encoder embeds the spectral-temporal information into a higher-dimensional hidden representation before progressively reducing it to retain the most relevant information. We adapted this architecture for classification by replacing the decoder with a classification head, which maps the learned feature representation to the target classes. In addition, we implemented an optional \gls{mha}-based pooling mechanism, inspired by the Pooling by Multihead Attention (PMA) approach proposed by \citet{lee_set_2018}, prior to the classification head. Unlike classification approaches that rely solely on the final hidden state, the attention pooling mechanism incorporates information from all RNN time steps. A learnable query attends to the temporal latent representations, assigning attention weights to the individual time steps and aggregating them into a single fixed-dimensional representation that is subsequently passed to the classification head. The architecture of an RNN model with \gls{mha} pooling is illustrated in Figure ~\ref{fig:lstmmhaarchitectureLS}.

\begin{figure}
    \includegraphics[width=1\columnwidth]{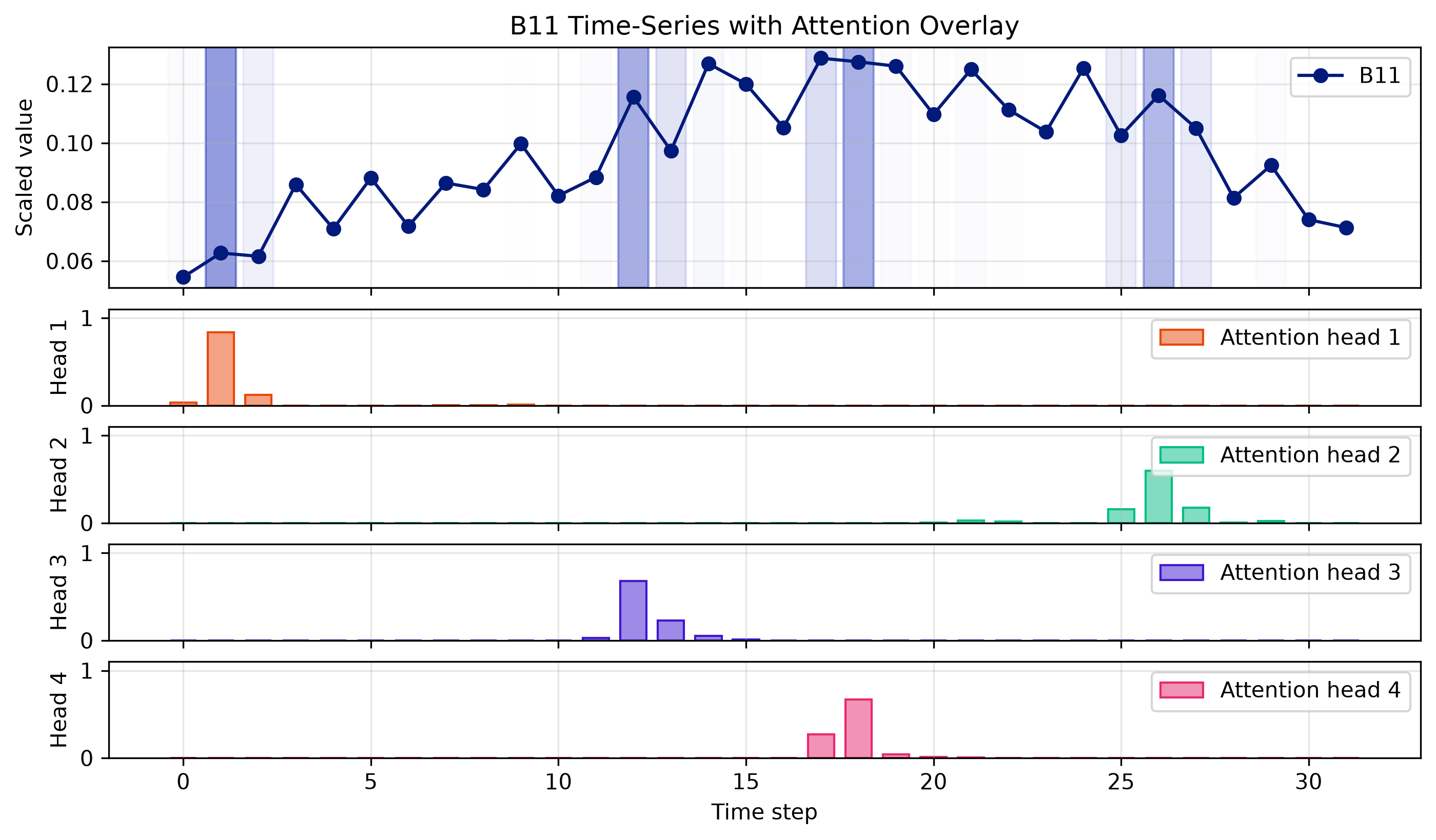}
    \caption{\textit{Multi-head attention weighting.} The top panel shows the temporal profile of spectral band 11 (SWIR 1), which is included for visualization purposes. The panels below show the attention weights assigned to each time step by the individual attention heads. The blue bars in the top panel represent the attention weights aggregated across all heads for each time step. The attention mechanism operates on the latent representations rather than directly on the spectral band values.}

    \label{fig:attentionmech}
\end{figure}

We incorporated \gls{mha} pooling to allow the model to weight the temporal representations produced by the recurrent encoder. Because time steps contribute unequally to classification, the \gls{mha} pooling enables the model to learn time steps containing particularly discriminative information. By using multiple attention heads, different heads can learn complementary weighting patterns \citep{vaswani_attention_2017}, potentially capturing information associated with distinct phenological periods. We therefore expected \gls{mha} pooling to produce a more informative latent-space representation without increasing the dimensionality of the latent representation. Consequently, we expected it to improve classification, particularly for classes whose spectral differences are most pronounced during specific periods of the growing season. This mechanism is schematically illustrated in Figure~\ref{fig:attentionmech}. For simplicity, attention is shown for a single spectral band, although in the model it is applied to the full embedded representation of all bands at each time step.

Because the time series are incomplete and irregularly spaced, we included the \gls{doy} as a scaled continuous input feature. This provided the recurrent models with explicit calendar-time information and relaxed the assumption that consecutive observations are equally spaced. Because each sequence covered a single calendar year and did not span the December–January transition, cyclic encoding of \gls{doy} was not required.

\subsection{Feature Importance}

Deep learning models are often considered black boxes because they provide limited insight into the factors driving their predictions \citep{samek_explainable_2017, hassija_interpreting_2024}. 
There are several attribution methods that assign a scalar contribution value to each input feature, aiming to quantify how strongly each feature influenced the network’s prediction \citep{ancona_gradient-based_2019}. These methods can make it difficult to determine whether observed inaccuracies originate from the model itself or from the attribution method \citep{sundararajan_axiomatic_2017}.
We selected \textit{\gls{ig}}, introduced by \citet{sundararajan_axiomatic_2017}, as the attribution method for this study. \gls{ig} is a widely used and theoretically grounded attribution method for explaining predictions of differentiable models \citep{jha_enhanced_2020,sanyal_discretized_2021}. It has also been shown to provide informative explanations for multivariate time-series models \citep{tronchin_translating_2024}.

Integrated Gradients is a gradient-based attribution method that assigns an attribution value to each input feature, that is, to each spectral band at each time step. For each input feature, \gls{ig} integrates the gradients of the model output along a linear interpolation path between a predefined baseline and the observed multivariate time series \citep{sundararajan_axiomatic_2017}. The resulting attribution values quantify how each feature contributes to the difference between the model output for the observed input and that for the baseline. Consequently, the resulting attribution matrix represents the contribution of each spectral-band-time-step combination to the prediction of the respective target class relative to the baseline input.

\begin{equation}
    \operatorname{IG}_i(x) ::= (x_i-x'_i) \times \int^1_{\alpha=0} = \frac{\partial F(x'+\alpha \times (x-x'))}{\partial x_i}d\alpha  
\end{equation}

Here $x$ denotes the input sample and $x'$ denotes the corresponding baseline input. The term $\frac{\partial F(x)}{\partial x_i}$ is the gradient of the model output $F(x)$ along the $i^{th}$ dimension. In practice, the integral is approximated using a Riemann sum with $m$ steps.  

\begin{equation}
\operatorname{IG}^{\mathrm{approx}}_i(x)::= (x_i - x'_i) \times \sum_{k=1}^{m} \frac{\partial F\left(x' + \frac{k}{m}\times(x - x')\right)}{\partial x_i} \times \frac{1}{m}
\end{equation}

The authors suggest comparing observations with a zero baseline to identify the input features most relevant to a model’s classification decision \citep{sundararajan_axiomatic_2017}. By applying IG to multiple observations and averaging the results, a more robust estimate of feature importance can be obtained. 

Because we focused on classes that were difficult to distinguish, we also used representative baselines derived from the respective comparison classes. These class-specific baselines allowed us to identify the spectral bands and time steps that contributed most strongly to distinguishing the observed class from the class used as the baseline.

\section{Methodology}
In this section, we describe the methodological framework used for forest classification and model interpretation. We first describe the data preprocessing steps, then the training and inference procedures for the recurrent neural network and ensemble models. We then introduce the evaluation procedure and the Integrated Gradients approach used to analyze the spectral and temporal relevance of the model inputs.
To support reproducibility, the code used in this study for data preprocessing, model development, evaluation, and Integrated Gradients analysis is publicly available on GitHub\footnote{GitHub repository: \url{https://github.com/Max25832/TreeClassXAI}}.
\subsection{Data Preprocessing} \label{chap31}

Many recent studies use multi-year satellite image time series for forest classification \citep{wegler_tree_2025, immitzer_first_2016, freudenberg_consistent_2025}. In contrast, we focused on single-season sequences because the objective of this study is to identify phenologically relevant time periods and spectral regions for separating similar forest structures, rather than to optimize an operational tree species mapping model.

The time series were therefore divided into yearly sequences that can be generalized across annual phenological developments. This reduced the risk of the models learning patterns specific to individual extreme events, such as the exceptionally hot and dry summer of 2018. Furthermore, this approach also increased the number of available training examples. Additionally, we excluded observations before 2017, because Sentinel-2B started in June 2017, leading to more observations per year \citep{freudenberg_sentinel-2_2025}. 
The train-test split was performed prior to temporal segmentation to avoid data leakage. This ensured that time series segments of the same pixel from different years were strictly assigned to either the training or the test dataset and did not appear in both datasets simultaneously. We used a standard 80:20 train-test split. To reduce class imbalance in the training data, the number of samples per class was limited to at most twice the sample size of the smallest class:
\begin{equation}
    n_c \leq 2 \cdot \min_{k \in \mathcal{C}} n_k \quad \forall c \in \mathcal{C}   
\end{equation}
where $n_c$ denotes the number of samples in class $c$, $\mathcal{C}$ denotes the set of all classes, $k$ is an index over the classes in $\mathcal{C}$, and $\min_{k \in \mathcal{C}} n_k$ denotes the sample size of the smallest class.
Downsampling was performed after the data were divided into yearly subsequences. This allowed us to retain a spatially and temporally diverse subset of the highly represented classes, as samples were randomly selected from different pixels and years, thereby increasing both spatial and temporal diversity.


\begin{table}[htbp]
  \caption{Final Sequences Sample Size}
  \label{tab:t2}
  \centering
  \small
  \setlength{\tabcolsep}{3pt}

  \begin{tabular*}{0.9 \columnwidth}
    {@{\extracolsep{\fill}}lcrr@{}}
    \toprule
    \multirow{2}{*}{Species}
      & \multirow{2}{*}{Description}
      & \multicolumn{2}{c}{Samples} \\
    \cmidrule(lr){3-4}
    & & Train & Test \\
    \midrule
    \textit{Fagus sylvatica} & pure stand & 13131 & 6828 \\
    \textit{Quercus spp.}   & pure stand & 7738  & 2244 \\
    \textit{Pinus sylvestris} & pure stand & 16702 & 32484 \\
    \textit{Picea abies} & pure stand      & 15989 & 19688 \\
    \textit{Picea abies} & mixed stand     & 16965 & 40511 \\
    \textit{Picea abies} & disturbed stand & 7767  & 2230 \\
                        & raw soil        & 15875 & 17992 \\
    \bottomrule
  \end{tabular*}
\end{table}

Because the deep learning models used sequential inputs, the original time-series structure could be retained. In contrast, the ensemble models required tabular input data. For these models, the time series were transformed into a wide-format representation, where each spectral band at each time step was represented by a separate predictor variable.
Using daily time steps would have resulted in $366 \text{ days} \times 10 \text{ spectral bands} = 3{,}660$ predictors. This would have produced a very high-dimensional dataset, increasing the risk of overfitting and leading to a large number of missing values, as the time series contained, on average, only approximately 24 observations per year. To reduce dimensionality and obtain a more compact representation, we aggregated the data to monthly median values. Remaining gaps were imputed by forward filling. As a result, the tabular dataset used for the ensemble models contained 120 predictors.

\subsection{Model Training and Inference}

Hyperparameter tuning for both ensemble models was performed on the training set using five-fold cross-validation and predefined tuning grids. The evaluated hyperparameters are provided in Appendix Table~\ref{tab:hyperparameter}.
For the deep learning models, the training set was further split into training and internal validation subsets using an 80:20 split. The internal validation subset was used for checkpoint selection during training, while the hold-out test set was excluded from training, validation, and hyperparameter tuning. After evaluation, test-set Macro-F1 was used to select the model configurations for the feature importance analysis described below. All RNN-based models were trained for 80 epochs using the Adam optimizer with a learning rate of $1 \times 10^{-3}$. 
For the loss function, we used a weighted combination of cross-entropy loss and a Macro-F1-based loss, with weights of 0.7 for cross-entropy loss and 0.3 for the Macro-F1-based loss. We additionally applied early stopping during training, using Macro-F1 on the validation set as the monitoring metric. 
Both recurrent architectures, GRU and LSTM, were trained with and without multi-head attention pooling. All RNN-based models used a bidirectional variant, in which the input sequence was processed in both forward and backward temporal directions. We tested bottleneck dimensions of 32, 64, and 128. The initial embedding dimension was set to four times the corresponding bottleneck dimension (see Appendix Table~\ref{tab:hyperparameter}).

After training, all models were evaluated on the held-out test set using accuracy, balanced accuracy, and Macro-F1. Balanced accuracy and Macro-F1 were included to account for class imbalance in the dataset.

\subsection{Application to a Mixed-Forest Area}

The dataset authors note that it is intended primarily for model training and is less suitable for independent validation \citep{freudenberg_sentinel-2_2025}. To account for this and to illustrate the practical application of the trained model, we visually selected an additional mixed-forest area in Thuringia from orthorectified aerial images \footnote{Orthophotos: \url{https://geoportal.thueringen.de/gdi-th/download-offene-geodaten/download-luftbilder-und-orthophotos}.}. We extracted Sentinel-2 time series for each pixel within the area using Google Earth Engine and classified them with the best-performing trained model. The resulting pixel-wise predictions were used to generate a species-composition map and to estimate the area assigned to each forest class. 
We used orthophotos from 2023 for visual confirmation and classified the years 2022 and 2023. For each pixel, we selected the year with the higher classification confidence to generate the final map.

\subsection{Feature Importance Analysis}

\begin{figure}[b]
     \begin{adjustbox}{max width=1\columnwidth}
    \input{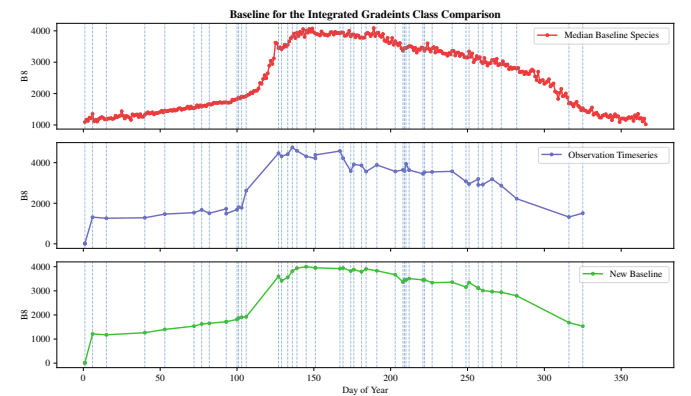}
    \end{adjustbox}   
    \caption{\textit{Class-specific baseline derivation for \gls{ig} attribution.} For each observation, a baseline with the same temporal structure was derived from the median time series of the comparison class. The example shows an oak observation (blue) and the corresponding beech baseline (green) for band 11, derived from the median of all beech observations (red) and matched to the observation time steps.}
    \label{fig:IG_baseline}
\end{figure}

To investigate model explainability, we applied \gls{ig} to quantify the contribution of individual spectral bands and time steps to each classification. For each of the main tree species, we initially computed attributions relative to a zero baseline. We randomly selected 2,000 multivariate time series per class from the test set and restricted the analysis to samples that were correctly classified by the respective model. This was done to ensure that the resulting attribution patterns reflected successful model decisions. The path integral in \gls{ig} was approximated using a Riemann sum with $m=30$ interpolation steps between the baseline and the observed input.
Furthermore, we compared observed time series against class-specific baselines derived from other classes to identify the spectral and temporal features that contributed most strongly to distinguishing the observed class from the respective baseline class. In this context, an individual input feature corresponds to the value of a single spectral band at a single time step. 
Because \gls{ig} requires the baseline and input to have the same shape,  we had to account for the varying sequence lengths. We therefore constructed a median baseline time series for each selected baseline species by calculating the median separately for each spectral band and day of year across all available observations. For each observed sequence, we then selected only those values from the median baseline that corresponded to the day of year present in that sequence, thereby ensuring temporal alignment between the input and its baseline. 
This process is illustrated for a single spectral band of one observation in Figure ~\ref{fig:IG_baseline}.

For the feature importance analysis, we selected the model configurations based on the highest test-set Macro-F1 scores. We did not compare models that differed only in embedding dimensions. In contrast, we separately analyzed models with and without multi-head attention pooling because these architectures aggregate the temporal representations differently and may therefore differ in attribution patterns. We compared all classes against the zero baseline. In addition, pairwise class-specific baseline comparisons were performed between \textit{pure Norway spruce} and \textit{mixed Norway spruce} and between \textit{oak} and \textit{beech}. Each pairwise comparison was conducted in both directions, with each class serving once as the observed class and once as the baseline class.   

\section{Results}

To evaluate model performance, we compared \gls{rnn} configurations with different recurrent architectures, input and bottleneck dimensions, and either \gls{mha} pooling or no pooling against the ensemble models. Performance was assessed primarily using the Macro-F1 score, supported by additional classification metrics. We further applied \gls{ig} to analyze the spectral and temporal patterns contributing to correct class predictions and conducted detailed class-comparison analyses for Norway spruce versus mixed Norway spruce and oak versus beech.

\subsection{Classification Performance}
Based on the Macro-F1 score, the GRU model with multi-head attention and a bottleneck dimension of 128 achieved the best overall performance, with a Macro-F1 score of 0.851. It additionally reached an accuracy of 0.854 and a balanced accuracy of 0.933. The GRU-128 model without attention performed similarly but achieved a slightly lower Macro-F1 score of 0.847, together with an accuracy of 0.853 and the same balanced accuracy of 0.933. Among the baseline models, XGBoost achieved the highest Macro-F1 score of 0.839 and an accuracy of 0.851, whereas the Random Forest showed the lowest overall performance. For both recurrent architectures, larger bottleneck dimensions generally corresponded to higher Macro-F1 scores, while the effect of multi-head attention was comparatively small. \gls{mha} did not improve the Macro-F1 performance of the \gls{lstm} models, but increased Macro-F1 by 0.004 for the \gls{gru} models. These results are shown in Table ~\ref{tab:model_comparison_all}.

\begin{table}[h]
    \caption{Comparison of Model Configurations}
    \label{tab:model_comparison_all}
    \begin{tabular}{lllccc}
        \toprule
        
        \multicolumn{3}{c}{Model}
        & \multicolumn{3}{c}{Performance metrics} \\
        
        \cmidrule(r{0.5em}){1-3}
        \cmidrule(l{0.5em}){4-6}
        
        RNN & Attention & BN Dim.
        & Acc
        & BA
        & Macro-F1 \\
        
        \midrule
        
        \multirow{6}{*}{LSTM}
        & \multirow{3}{*}{MHA}
        & 32  & 0.832 & 0.909 & 0.825 \\
        & & 64  & 0.827 & 0.920 & 0.814 \\
        & & 128 & 0.848 & 0.930 & 0.843 \\
        
        \cmidrule(r{0.8em}){2-3}
        \cmidrule(l{0.8em}){4-6}
        
        & \multirow{3}{*}{None}
        & 32  & 0.821 & 0.912 & 0.813 \\
        & & 64  & 0.830 & 0.923 & 0.830 \\
        & & 128 & 0.850 & 0.931 & 0.843 \\
        
        \midrule
        
        \multirow{6}{*}{GRU}
        & \multirow{3}{*}{MHA}
        & 32  & 0.832 & 0.917 & 0.821 \\
        & & 64  & 0.840 & 0.924 & 0.835 \\
        & & 128 & \textbf{0.854} & \textbf{0.933} & \textbf{0.851} \\
        
        \cmidrule(r{0.8em}){2-3}
        \cmidrule(l{0.8em}){4-6}
        
        & \multirow{3}{*}{None}
        & 32  & 0.825 & 0.915 & 0.813 \\
        & & 64  & 0.828 & 0.925 & 0.832 \\
        & & 128 & 0.853 & \textbf{0.933} & 0.847 \\
        
        \midrule
        
        \multicolumn{3}{c}{\gls*{xgb}} & 0.851 & 0.925 & 0.839 \\
        \midrule
        \multicolumn{3}{c}{RF} & 0.792 & 0.890 & 0.748 \\
        
        \bottomrule
    \end{tabular}
     \par\smallskip
    \begin{minipage}{\columnwidth}
        \footnotesize
        * BN Dim. = Model Bottleneck Dimension; Acc = accuracy; BA = balanced accuracy.
    \end{minipage}
\end{table}

\begin{figure*}
    \centering
    \begin{adjustbox}{max width=0.9\linewidth}
            \input{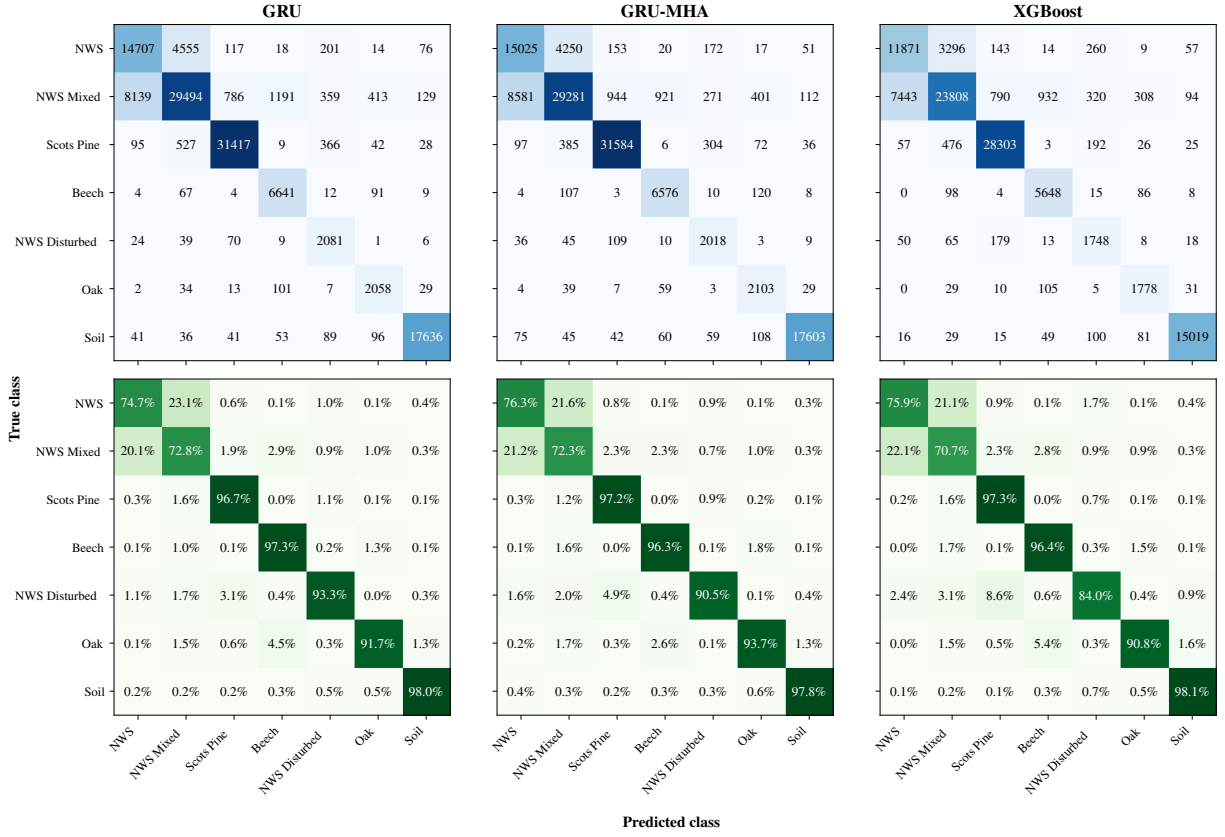}
    \end{adjustbox}

    \caption{\textit{Confusion matrices for the two best-performing RNN models, GRU with and without attention, and the XGBoost model}
}
    \label{fig:confusionmatrices}
\end{figure*}

The confusion matrices in Figure~\ref{fig:confusionmatrices} revealed that the main classification challenge was distinguishing Norway spruce from mixed Norway spruce. Across all models, approximately one-fifth of the samples in each class were confused with the other class. The green confusion matrices are normalized by true class and report the proportion of samples in percent; therefore, the diagonal entries correspond to class-specific sensitivity. For the GRU-MHA model, 21.6\% of Norway spruce samples were classified as mixed Norway spruce, while 21.2\% of mixed Norway spruce samples were assigned to Norway spruce. Accordingly, class-specific sensitivity was comparatively low for Norway spruce (76.3\%) and mixed Norway spruce (72.3\%).
The remaining classes achieved substantially higher sensitivities. The GRU-MHA model reached 97.2\% for Scots pine, 96.3\% for beech, 90.5\% for disturbed Norway spruce, 93.7\% for oak, and 97.8\% for soil. Compared with the corresponding GRU model without MHA, the addition of MHA reduced the proportion of oak samples misclassified as beech from 4.5\% to 2.6\%, while the reverse confusion increased slightly from 1.3\% to 1.8\%. This indicates that MHA produced only small, class-specific changes in the confusion patterns. The mutual confusion between pure and mixed Norway spruce remained the dominant source of classification error.

The class-specific evaluation metrics support the patterns observed in the confusion matrices. Norway spruce achieved the lowest F1 scores, ranging from 0.614 to 0.691. The GRU-128 model with multi-head attention achieved the highest F1 score (0.691) and sensitivity (0.763), while the highest specificity was 0.919. For mixed Norway spruce, F1 scores ranged from 0.697 to 0.784, with two recurrent models tying for the highest score. The maximum sensitivity and specificity were 0.729 and 0.943, respectively. For both classes, specificity exceeded sensitivity, indicating that false-negative rates were higher than false-positive rates.
The remaining classes achieved higher F1 scores. The best scores were obtained by the GRU-128 model with multi-head attention for disturbed areas (0.797), the GRU-128 model without attention for Scots pine (0.968), and the LSTM-128 model without attention for beech (0.909). XGBoost achieved the highest scores for oak (0.836) and soil (0.983), although the recurrent models performed similarly. Overall, soil and Scots pine had the highest class-specific F1 scores, whereas Norway spruce had the lowest. Random Forest generally underperformed compared with XGBoost and the recurrent models. These results are presented in detail in Table~\ref{tab:model_comparison_tree_spec}.

\begin{table*}
    \centering
    \caption{Per-class comparison of model configurations}
    \label{tab:model_comparison_tree_spec}
    \footnotesize

    \begin{tabular*}{0.9\textwidth}{@{\extracolsep{\fill}}llcccccc@{}}
        \toprule
    
        \multirow{2}{*}{Tree species}
        & \multirow{2}{*}{Metric}
        & \multirow{2}{*}{\textbf{RF}}
        & \multirow{2}{*}{\textbf{\gls*{xgb}}}
        & \multicolumn{2}{c}{\textbf{LSTM-128}}
        & \multicolumn{2}{c}{\textbf{GRU-128}}
        \\
        \cmidrule(lr{0.4em}){5-6}
        \cmidrule(l{0.4em}){7-8}
        
        & & & & MHA & No Att. & MHA & No Att. \\
        
        \toprule
        \multirow{3}{*}{Norway spruce}
        & F1          & 0.614 & 0.677 & 0.677 & 0.684 & \textbf{0.691} & 0.689 \\
        & Sensitivity & 0.704 & 0.759 & 0.731 & 0.742 & \textbf{0.763} & 0.747 \\
        & Specificity & 0.895 & 0.914 & 0.917 & 0.918 & 0.914 & \textbf{0.919} \\
        \midrule
        \addlinespace[0.25em]
        
        \multirow{3}{*}{Norway spruce mixed}
        & F1          & 0.697 & 0.774 & 0.775 & 0.782 & \textbf{0.784} & \textbf{0.784}\\
        & Sensitivity & 0.610 & 0.707 & 0.717 & \textbf{0.729} & 0.723 & 0.728 \\
        & Specificity & 0.932 & \textbf{0.943} & 0.934 & 0.933 & 0.940 & 0.935 \\
        
        \midrule
        \addlinespace[0.15em]
        
        \multirow{3}{*}{Scots pine}
        & F1          & 0.935 & 0.967 & 0.966 & 0.962 & 0.967 & \textbf{0.968} \\
        & Sensitivity & 0.935 & 0.973 & \textbf{0.973} & 0.964 & 0.972 & 0.967 \\
        & Specificity & 0.975 & 0.985 & 0.985 & 0.985 & 0.986 & \textbf{0.988} \\
        
        \midrule
        \addlinespace[0.15em]
        
        \multirow{3}{*}{Beech}
        & F1          & 0.814 & 0.895 & 0.891 & \textbf{0.909} & 0.908 & 0.894 \\
        & Sensitivity & 0.905 & 0.964 & 0.966 & 0.949 & 0.963 & \textbf{0.973} \\
        & Specificity & 0.981 & 0.989 & 0.988 & \textbf{0.992} & 0.991 & 0.988 \\
        
        \midrule
        \addlinespace[0.15em]
        
        \multirow{3}{*}{Oak}
        & F1          & 0.646 & \textbf{0.836} & 0.835 & 0.809 & 0.830 & 0.830 \\
        & Sensitivity & 0.825 & 0.908 & 0.924 & \textbf{0.938} & 0.937 & 0.917 \\
        & Specificity & 0.986 & \textbf{0.995} & 0.995 & 0.993 & 0.994 & 0.995 \\
        
        \midrule
        \addlinespace[0.15em]
        
        \multirow{3}{*}{Soil}
        & F1          & 0.971 & \textbf{0.983} & 0.982 & 0.980 & 0.982 & 0.982 \\
        & Sensitivity & 0.963 & \textbf{0.981} & 0.980 & 0.978 & 0.978 & 0.980 \\
        & Specificity & 0.996 & 0.997 & 0.997 & 0.997 & \textbf{0.998} & 0.997 \\
        
        \midrule
        \addlinespace[0.15em]
        
        \multirow{3}{*}{Disturbed}
        & F1          & 0.556 & 0.741 & 0.776 & 0.776 & \textbf{0.797} & 0.779 \\
        & Sensitivity & 0.772 & 0.840 & 0.913 & 0.921 & 0.905 & \textbf{0.933} \\
        & Specificity & 0.979 & 0.991 & 0.992 & 0.992 & \textbf{0.993} & 0.991 \\
        
        \bottomrule
    \end{tabular*}
\end{table*}

\newpage
\subsection{Performance in a Mixed-Forest Area}

\begin{figure*}
    \centering

    \begin{subfigure}[t]{0.70\textwidth}
        \vspace{0pt}
        \centering
        \includegraphics[width=\linewidth]{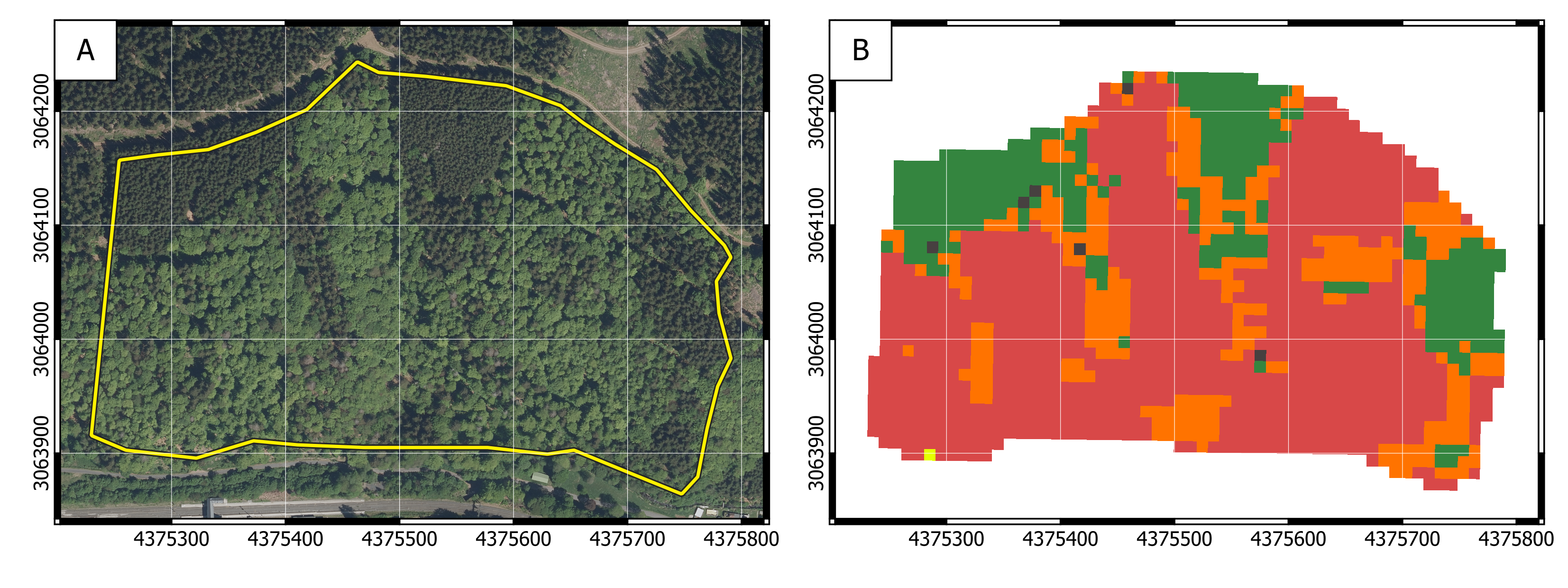}
        \caption{Predicted forest classes}
        \label{fig:classificationMap}
    \end{subfigure}%
    \hfill
    \begin{subfigure}[t]{0.28\textwidth}
        \vspace{0pt}
        \centering
        \begin{adjustbox}{max width=\linewidth}
\begin{tikzpicture}[x=1pt,y=1pt]
\definecolor{fillColor}{RGB}{255,255,255}
\path[use as bounding box,fill=fillColor,fill opacity=0.00] (0,0) rectangle (433.62,361.35);
\begin{scope}
\path[clip] (  0.00,  0.00) rectangle (433.62,361.35);
\definecolor{drawColor}{RGB}{255,255,255}
\definecolor{fillColor}{RGB}{255,255,255}

\path[draw=drawColor,line width= 0.6pt,line join=round,line cap=round,fill=fillColor] (  0.00,  0.00) rectangle (433.62,361.35);
\end{scope}
\begin{scope}
\path[clip] (  0.00,  0.00) rectangle (433.62,361.35);
\definecolor{fillColor}{RGB}{255,255,255}

\path[fill=fillColor] ( 47.99, 21.60) rectangle (428.62,331.35);
\definecolor{drawColor}{RGB}{216,72,72}
\definecolor{fillColor}{RGB}{216,72,72}

\path[draw=drawColor,line width= 0.6pt,fill=fillColor,fill opacity=0.70] ( 58.97, 35.68) rectangle (124.85,317.27);
\definecolor{drawColor}{RGB}{71,64,64}
\definecolor{fillColor}{RGB}{71,64,64}

\path[draw=drawColor,line width= 0.6pt,fill=fillColor,fill opacity=0.70] (132.17, 35.68) rectangle (198.05, 36.90);
\definecolor{drawColor}{RGB}{52,133,63}
\definecolor{fillColor}{RGB}{52,133,63}

\path[draw=drawColor,line width= 0.6pt,fill=fillColor,fill opacity=0.70] (205.37, 35.68) rectangle (271.24, 92.58);
\definecolor{drawColor}{RGB}{255,115,0}
\definecolor{fillColor}{RGB}{255,115,0}

\path[draw=drawColor,line width= 0.6pt,fill=fillColor,fill opacity=0.70] (278.56, 35.68) rectangle (344.44, 94.42);
\definecolor{drawColor}{RGB}{229,255,0}
\definecolor{fillColor}{RGB}{229,255,0}

\path[draw=drawColor,line width= 0.6pt,fill=fillColor,fill opacity=0.70] (351.76, 35.68) rectangle (417.64, 35.92);
\definecolor{drawColor}{RGB}{0,0,0}

\node[text=drawColor,anchor=base,inner sep=0pt, outer sep=0pt, scale=  1.20] at ( 91.91,329.67) {11.48 ha};

\node[text=drawColor,anchor=base,inner sep=0pt, outer sep=0pt, scale=  1.20] at (165.11, 49.30) {0.05 ha};

\node[text=drawColor,anchor=base,inner sep=0pt, outer sep=0pt, scale=  1.20] at (238.31,104.98) {2.32 ha};

\node[text=drawColor,anchor=base,inner sep=0pt, outer sep=0pt, scale=  1.20] at (311.50,106.82) {2.395 ha};

\node[text=drawColor,anchor=base,inner sep=0pt, outer sep=0pt, scale=  1.20] at (384.70, 48.32) {0.01 ha};
\end{scope}
\begin{scope}
\path[clip] (  0.00,  0.00) rectangle (433.62,361.35);
\definecolor{drawColor}{RGB}{0,0,0}

\path[draw=drawColor,line width= 0.6pt,line join=round] ( 47.99, 21.60) --
	( 47.99,331.35);
\end{scope}
\begin{scope}
\path[clip] (  0.00,  0.00) rectangle (433.62,361.35);
\definecolor{drawColor}{RGB}{0,0,0}

\node[text=drawColor,anchor=base east,inner sep=0pt, outer sep=0pt, scale=  1.20] at ( 42.59, 31.54) {0};

\node[text=drawColor,anchor=base east,inner sep=0pt, outer sep=0pt, scale=  1.20] at ( 42.59,105.13) {300};

\node[text=drawColor,anchor=base east,inner sep=0pt, outer sep=0pt, scale=  1.20] at ( 42.59,178.72) {600};

\node[text=drawColor,anchor=base east,inner sep=0pt, outer sep=0pt, scale=  1.20] at ( 42.59,252.31) {900};

\node[text=drawColor,anchor=base east,inner sep=0pt, outer sep=0pt, scale=  1.20] at ( 42.59,325.89) {1200};
\end{scope}
\begin{scope}
\path[clip] (  0.00,  0.00) rectangle (433.62,361.35);
\definecolor{drawColor}{gray}{0.20}

\path[draw=drawColor,line width= 0.6pt,line join=round] ( 44.99, 35.68) --
	( 47.99, 35.68);

\path[draw=drawColor,line width= 0.6pt,line join=round] ( 44.99,109.26) --
	( 47.99,109.26);

\path[draw=drawColor,line width= 0.6pt,line join=round] ( 44.99,182.85) --
	( 47.99,182.85);

\path[draw=drawColor,line width= 0.6pt,line join=round] ( 44.99,256.44) --
	( 47.99,256.44);

\path[draw=drawColor,line width= 0.6pt,line join=round] ( 44.99,330.03) --
	( 47.99,330.03);
\end{scope}
\begin{scope}
\path[clip] (  0.00,  0.00) rectangle (433.62,361.35);
\definecolor{drawColor}{RGB}{0,0,0}

\path[draw=drawColor,line width= 0.6pt,line join=round] ( 47.99, 21.60) --
	(428.62, 21.60);
\end{scope}
\begin{scope}
\path[clip] (  0.00,  0.00) rectangle (433.62,361.35);
\definecolor{drawColor}{gray}{0.20}

\path[draw=drawColor,line width= 0.6pt,line join=round] ( 91.91, 18.60) --
	( 91.91, 21.60);

\path[draw=drawColor,line width= 0.6pt,line join=round] (165.11, 18.60) --
	(165.11, 21.60);

\path[draw=drawColor,line width= 0.6pt,line join=round] (238.31, 18.60) --
	(238.31, 21.60);

\path[draw=drawColor,line width= 0.6pt,line join=round] (311.50, 18.60) --
	(311.50, 21.60);

\path[draw=drawColor,line width= 0.6pt,line join=round] (384.70, 18.60) --
	(384.70, 21.60);
\end{scope}
\begin{scope}
\path[clip] (  0.00,  0.00) rectangle (433.62,361.35);
\definecolor{drawColor}{RGB}{0,0,0}

\node[text=drawColor,anchor=base,inner sep=0pt, outer sep=0pt, scale=  1.20] at (238.31,  7.33) {Class};
\end{scope}
\begin{scope}
\path[clip] (  0.00,  0.00) rectangle (433.62,361.35);
\definecolor{drawColor}{RGB}{0,0,0}

\node[text=drawColor,rotate= 90.00,anchor=base,inner sep=0pt, outer sep=0pt, scale=  1.20] at ( 13.26,176.47) {Pixel Count};
\end{scope}
\begin{scope}
\path[clip] (  0.00,  0.00) rectangle (433.62,361.35);
\definecolor{fillColor}{RGB}{255,255,255}

\path[fill=fillColor] (201.49,223.42) rectangle (421.01,331.35);
\end{scope}
\begin{scope}
\path[clip] (  0.00,  0.00) rectangle (433.62,361.35);
\definecolor{drawColor}{RGB}{0,0,0}

\node[text=drawColor,anchor=base west,inner sep=0pt, outer sep=0pt, scale=  2.00] at (207.49,309.63) {Class};
\end{scope}
\begin{scope}
\path[clip] (  0.00,  0.00) rectangle (433.62,361.35);
\definecolor{drawColor}{RGB}{216,72,72}
\definecolor{fillColor}{RGB}{216,72,72}

\path[draw=drawColor,line width= 0.6pt,fill=fillColor,fill opacity=0.70] (208.26,288.01) rectangle (221.17,300.91);
\end{scope}
\begin{scope}
\path[clip] (  0.00,  0.00) rectangle (433.62,361.35);
\definecolor{drawColor}{RGB}{71,64,64}
\definecolor{fillColor}{RGB}{71,64,64}

\path[draw=drawColor,line width= 0.6pt,fill=fillColor,fill opacity=0.70] (208.26,273.56) rectangle (221.17,286.46);
\end{scope}
\begin{scope}
\path[clip] (  0.00,  0.00) rectangle (433.62,361.35);
\definecolor{drawColor}{RGB}{52,133,63}
\definecolor{fillColor}{RGB}{52,133,63}

\path[draw=drawColor,line width= 0.6pt,fill=fillColor,fill opacity=0.70] (208.26,259.10) rectangle (221.17,272.00);
\end{scope}
\begin{scope}
\path[clip] (  0.00,  0.00) rectangle (433.62,361.35);
\definecolor{drawColor}{RGB}{255,115,0}
\definecolor{fillColor}{RGB}{255,115,0}

\path[draw=drawColor,line width= 0.6pt,fill=fillColor,fill opacity=0.70] (208.26,244.65) rectangle (221.17,257.55);
\end{scope}
\begin{scope}
\path[clip] (  0.00,  0.00) rectangle (433.62,361.35);
\definecolor{drawColor}{RGB}{229,255,0}
\definecolor{fillColor}{RGB}{229,255,0}

\path[draw=drawColor,line width= 0.6pt,fill=fillColor,fill opacity=0.70] (208.26,230.19) rectangle (221.17,243.10);
\end{scope}
\begin{scope}
\path[clip] (  0.00,  0.00) rectangle (433.62,361.35);
\definecolor{drawColor}{RGB}{0,0,0}

\node[text=drawColor,anchor=base west,inner sep=0pt, outer sep=0pt, scale=  1.60] at (227.94,288.95) {Beech};
\end{scope}
\begin{scope}
\path[clip] (  0.00,  0.00) rectangle (433.62,361.35);
\definecolor{drawColor}{RGB}{0,0,0}

\node[text=drawColor,anchor=base west,inner sep=0pt, outer sep=0pt, scale=  1.60] at (227.94,274.50) {Norway Spruce: Disturbed};
\end{scope}
\begin{scope}
\path[clip] (  0.00,  0.00) rectangle (433.62,361.35);
\definecolor{drawColor}{RGB}{0,0,0}

\node[text=drawColor,anchor=base west,inner sep=0pt, outer sep=0pt, scale=  1.60] at (227.94,260.04) {Norway Spruce: Healthy};
\end{scope}
\begin{scope}
\path[clip] (  0.00,  0.00) rectangle (433.62,361.35);
\definecolor{drawColor}{RGB}{0,0,0}

\node[text=drawColor,anchor=base west,inner sep=0pt, outer sep=0pt, scale=  1.60] at (227.94,245.59) {Norway Spruce: Mixed};
\end{scope}
\begin{scope}
\path[clip] (  0.00,  0.00) rectangle (433.62,361.35);
\definecolor{drawColor}{RGB}{0,0,0}

\node[text=drawColor,anchor=base west,inner sep=0pt, outer sep=0pt, scale=  1.60] at (227.94,231.13) {soil};
\end{scope}
\end{tikzpicture}
        \end{adjustbox}
        \caption{Classification statistics}
        \label{fig:map_stats}
    \end{subfigure}

    \caption{Map with distribution statistics of predicted forest classes.}
    \label{fig:ClassificationMapWStats}
\end{figure*}
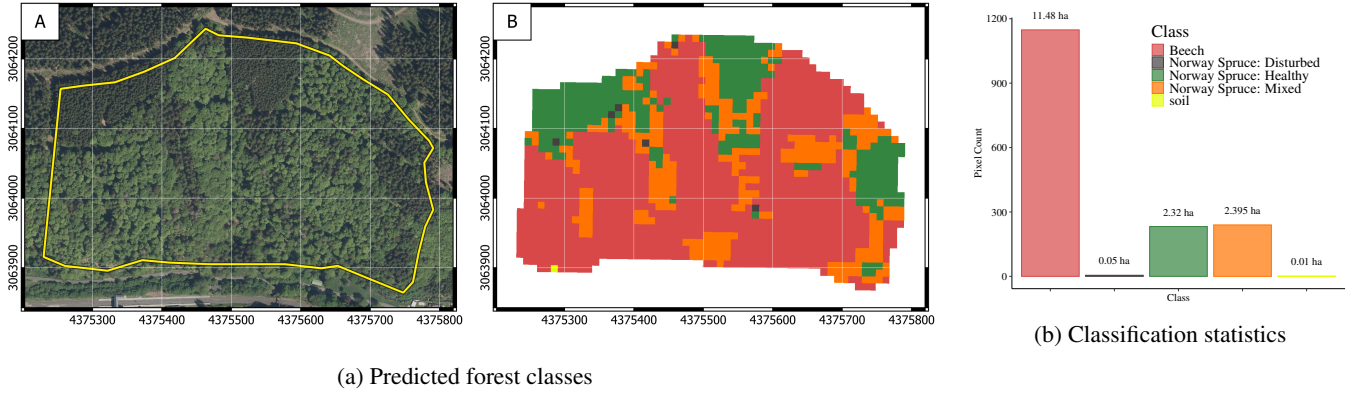

To additionally assess model performance in a mixed-forest area, we applied the GRU-MHA-128 model to single-year time series from a randomly selected site in Thuringia. The resulting classification map is shown in Fig.~\ref{fig:ClassificationMapWStats}. 

The model classified 11.48 ha as beech, 2.32 ha as pure Norway spruce, 2.40 ha as mixed Norway spruce, 0.05 ha as disturbed forest, and 0.01 ha as open soil. Because pixel-level labels were not available, we used an orthorectified aerial image as a qualitative visual reference. The visual comparison indicated reasonable agreement between the predicted classes and the aerial image, providing an additional qualitative assessment of model performance outside the original dataset.

\subsection{Model Explainability}

To improve the explainability of the deep learning model, we applied \gls{ig} in two ways. First, individual class attributions were calculated using a zero baseline. Second, we directly compared Norway spruce with mixed Norway spruce and oak with beech by using the respective opposing class as the baseline. The results of these analyses are presented in the following sections.

\subsubsection{Class-Specific Integrated Gradients}
Hereafter, spectral bands are referred to by their band identifiers, as defined in Table~\ref{tab:sentinel2_bands}.

\begin{table}[H]
    \centering
    \caption{Sentinel-2 Level-2A bands used in this study.}
        \label{tab:sentinel2_bands}
      \small
    \begin{tabular}{ccc}\toprule
         Band Identifier &  Band Name & Central Wavelength (nm)\\\midrule
         B2&  Blue& 492.7\\
         B3& Green &559.8 \\
         B4&  Red& 664.6\\
         B5&  Red Edge 1 &704.1 \\
         B6&  Red Edge 1& 740.5\\
         B7&  Red Edge 1& 782.8\\
         B8&  NIR & 832.8\\
         B8A&  Narrow NIR& 864.7\\
         B11& SWIR 1 & 1613.7\\
         B12&  SWIR 2 & 2202.4\\ 
         \bottomrule
    \end{tabular}
    \smallskip
        \begin{minipage}{\linewidth}
            \centering
            \footnotesize
            * NIR = near infrared; SWIR = short-wave infrared.
        \end{minipage}
\end{table}

The feature attribution analysis revealed distinct temporal and spectral patterns among the classes. For beech, attribution values peaked around \gls{doy} 120 in bands B3, B5, B6, and B11. Oak showed a similar temporal pattern, but with lower attribution values in bands B3–B6 and an earlier peak in B12, approximately between \gls{doy} 110 and 125.
For Scots pine, the \gls{ig} analysis showed high B11 attribution values between approximately \gls{doy} 130 and 190, whereas B6 exhibited moderate attribution values throughout the year. Pure Norway spruce and mixed Norway spruce showed similar attribution patterns, with high values in B6 around \gls{doy} 150. For mixed Norway spruce, the B6 attributions were concentrated between approximately \gls{doy} 140 and 175. In contrast, for pure Norway spruce, they extended from approximately \gls{doy} 120 to 220. Both Norway spruce and Scots pine showed elevated attribution values between approximately \gls{doy} 100 and 180. However, Scots pine also exhibited elevated attributions toward the end of the year. The disturbed Norway spruce class showed no pronounced temporal attribution pattern. Attribution values were broadly similar across bands, except for B2, B7, and B12.

\begin{figure*}
    \centering
     \begin{adjustbox}{max width=0.9\textwidth}
            \input{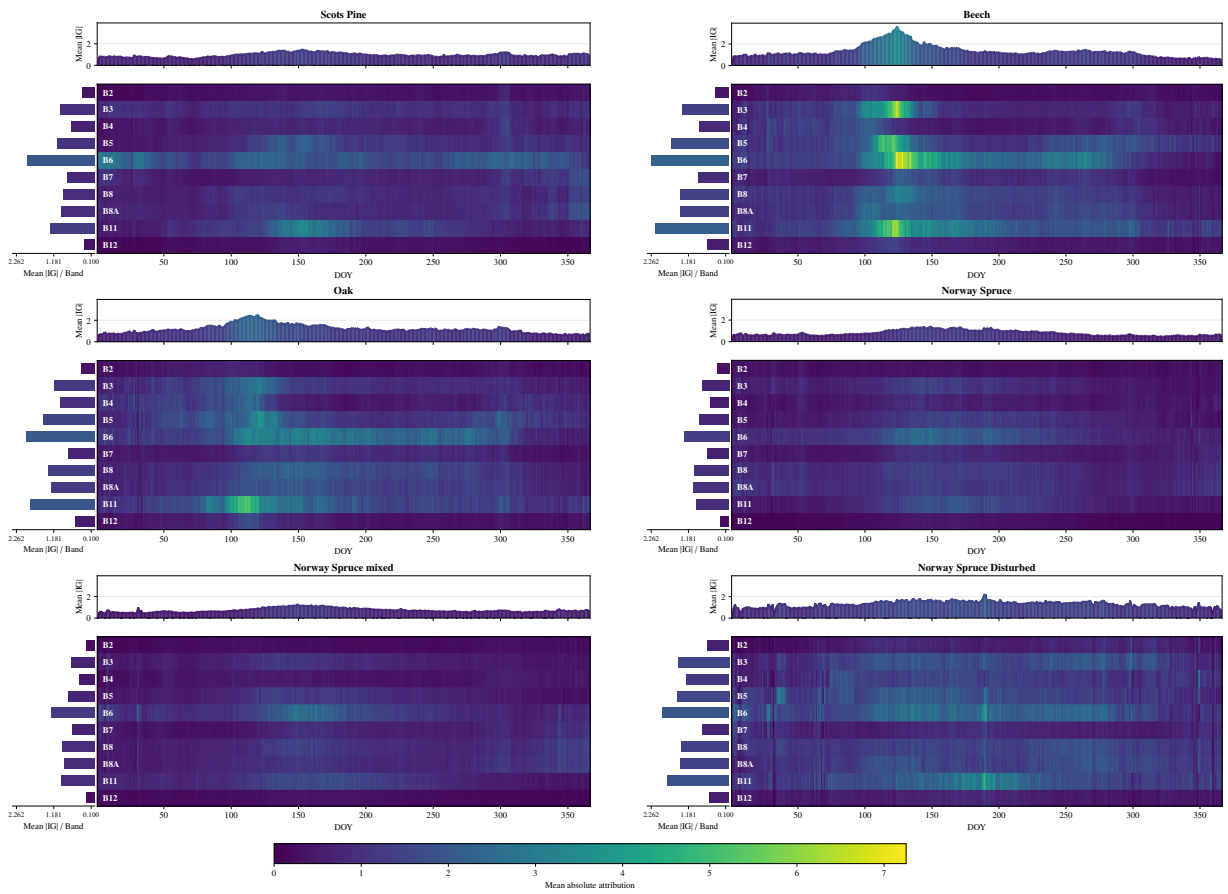}
            \end{adjustbox}
            \caption{Integrated Gradients attribution results for the six investigated classes against a zero-baseline.}
    \label{fig:attributions_0BL}
\end{figure*}

Overall, the \gls{ig} analysis revealed consistent differences in attribution magnitude among the spectral bands. Across classes, B2 and B12 generally showed the lowest attribution values, whereas B6 and B11 showed the highest attributions for correct species classifications. The results are visualized in Figure ~\ref{fig:attributions_0BL}.

\subsubsection{Integrated Gradients for Class Comparison}

The \gls{ig} analysis focused on the distinction between pure and mixed Norway spruce stands and between beech and oak. Because beech and oak were classified with comparatively high performance, the following analysis focuses on the attribution patterns for pure and mixed Norway spruce, while also examining differences between beech and oak.

\begin{figure}[H]
     \begin{adjustbox}{max width=1\columnwidth}
        \input{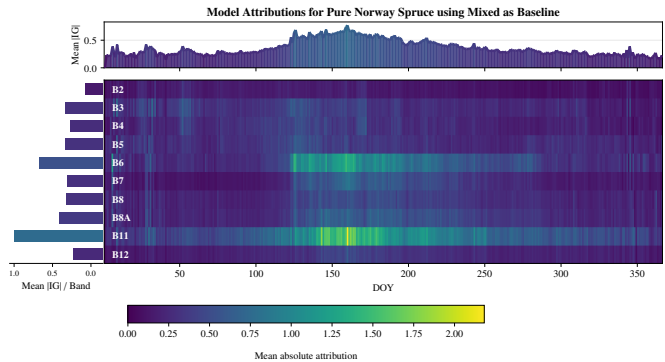}
            \end{adjustbox}
    \caption{\textit{Attribution map for the separation of pure and mixed Norway spruce.}
    }
    \label{fig:nwsvsmwsMixedBL}
\end{figure}

In the comparison between pure and mixed Norway spruce, B6 and B11 showed the highest overall attribution values. Most other bands exhibited comparable cumulative attributions across the year, whereas B2 and B12 consistently showed lower values. B11 exhibited elevated attributions throughout much of the year, with a pronounced peak between approximately \gls{doy} 140 and 180. Attributions for B6 increased after approximately \gls{doy} 125 and remained elevated until around \gls{doy} 290. The visible bands B2, B3, and B4 and the red-edge band B5 showed comparatively low but persistent attributions throughout the year. Bands B7, B8, and B8A followed a temporal pattern similar to that of B6, although their attribution magnitudes were generally lower. Overall, the highest attribution values for distinguishing pure from mixed Norway spruce occurred between approximately \gls{doy} 125 and 225. This comparison is visualized in Figure ~\ref{fig:nwsvsmwsMixedBL}.

\begin{figure}[H]
        \begin{adjustbox}{max width=1\columnwidth}
        \input{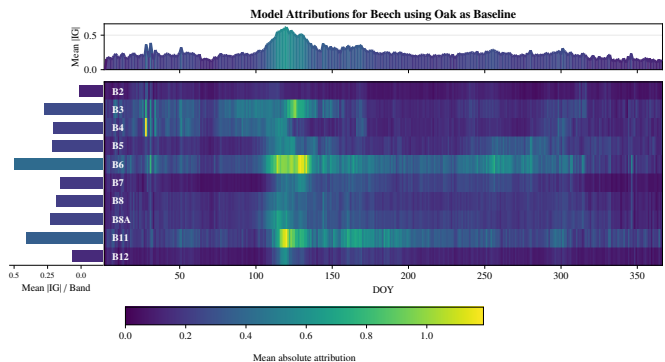}
            \end{adjustbox}
    \caption{\textit{Attribution Map for the separation of beech and oak}
    }
    \label{fig:beech_vs_oak}
\end{figure}

For beech versus oak, B3, B5, and B6 showed attribution patterns similar to those observed for beech relative to the zero baseline. However, B6 exhibited elevated attributions over a broader period, approximately from \gls{doy} 110 to 140. The highest attribution values were observed for B6 and B11, particularly between approximately \gls{doy} 110 and 150. This comparison is visualized in Figure ~\ref{fig:beech_vs_oak}.

\section{Discussion}

This study demonstrates the value of recurrent neural networks and attribution-based analysis for classifying and interpreting annual multispectral time series. The RNN models outperformed XGBoost, highlighting their ability to capture temporal dependencies in the data. Combined with Integrated Gradients, they also enabled a more fine-grained interpretation of relevant spectral bands and time periods than feature-importance measures from tree-based models, which are typically aggregated over broader temporal intervals. In addition, the class-comparison attribution analysis provided insights beyond the conventional zero-baseline approach by revealing shifts in the relative importance of individual spectral bands. These findings show that directly comparing classes can improve the interpretation of spectral and temporal differences between closely related forest classes.

\subsection{Comparative Model Performance}

The class-specific results indicate that classification performance was primarily determined by the separability of the spectral-temporal classes rather than by the architecture of the model. The main source of error across all configurations was the mutual confusion between pure and mixed Norway spruce, with approximately one-fifth of the observations misclassified between these two classes. This persistent confusion is likely due to the strong overlap in multispectral and temporal characteristics between the two classes. Pure stands tend to occupy more distinct regions of spectral-temporal feature space, whereas mixed pixels are strongly influenced by their species composition and form a gradual spectral continuum. The difficulty of separating these classes therefore appears to be constrained not only by model capacity but also by the information content of the Sentinel-2 time series and the definition of the classes themselves. 
In contrast, Scots pine, beech, and soil were consistently classified with high accuracy across model architectures, further supporting the interpretation that class separability was the dominant factor determining classification success. Disturbed Norway spruce was also separated more clearly from healthy spruce than pure and mixed healthy spruce were from each other. Disturbance alters canopy condition and therefore produces stronger deviations in spectral-temporal trajectories, which may produce a more distinctive spectral-temporal signal than differences in species composition within otherwise healthy spruce stands.

Despite these class-dependent limitations, the recurrent models achieved the highest overall classification performance, with the GRU using multi-head attention and a bottleneck dimension of 128 reaching the highest Macro-F1 score of 0.851, while Random Forest consistently performed worst. However, the improvement over XGBoost was relatively small. This suggests that, although explicit sequence modeling provides an advantage for annual multispectral time-series classification, much of the relevant temporal information can also be captured through engineered or temporally aggregated features. 
The benefit of recurrent modeling was particularly evident for disturbed Norway spruce, where the recurrent models achieved substantially higher F1 scores and sensitivity than XGBoost, indicating that explicit modeling of temporal dependencies may be especially useful for classes characterized by changes in canopy condition over time. In contrast, improvements for most healthy tree-species classes were small, and XGBoost remained competitive or even performed best for some well-separated classes.

Larger bottleneck dimensions generally improved the performance of both recurrent architectures, suggesting that sufficient latent capacity may be more important than the choice between LSTM and GRU or the inclusion of attention. Although larger bottleneck dimensions may improve classification by preserving more information, they also increase the risk of overfitting. This trade-off should be considered when selecting the latent representation size.
The funnel-shaped encoder progressively compresses the hidden representation and may therefore encourage the model to retain the most relevant information. However, the present experiments cannot determine whether the funnel architecture itself is beneficial or whether the observed gains primarily result from the increased latent dimensionality and overall model capacity. A direct comparison with a non-funneling architecture of similar size should be considered in future work.

\gls{mha} had only a limited effect on predictive performance. It produced a marginal improvement for the GRU-128 model and did not improve the Macro-F1 score of the LSTM-128 model, suggesting that the recurrent hidden states already encoded most of the temporal information required for classification. Nevertheless, \gls{mha} pooling improved the Macro-F1 score across all GRU configurations, with somewhat larger gains for models with smaller bottleneck dimensions, and produced minor class-specific improvements. Attention may therefore compensate for limited latent capacity by suppressing less informative temporal representations. However, its main value in this study lies in improving the temporal interpretation of the model by emphasizing relevant time steps and making their contribution more apparent in the attribution analysis.

Overall, the performance gains over XGBoost were class-dependent and moderate. Latent-space size had a clearer influence on performance than \gls{mha}, while the remaining errors were driven mainly by limited spectral-temporal separability rather than by the selected model architecture. The main advantage of the RNNs, however, lies in their temporally structured representation, which supports a more detailed interpretation of phenologically relevant periods.

\subsection{Explainability}

\begin{figure}[H]
    \centering
    \includegraphics[width=1\columnwidth]{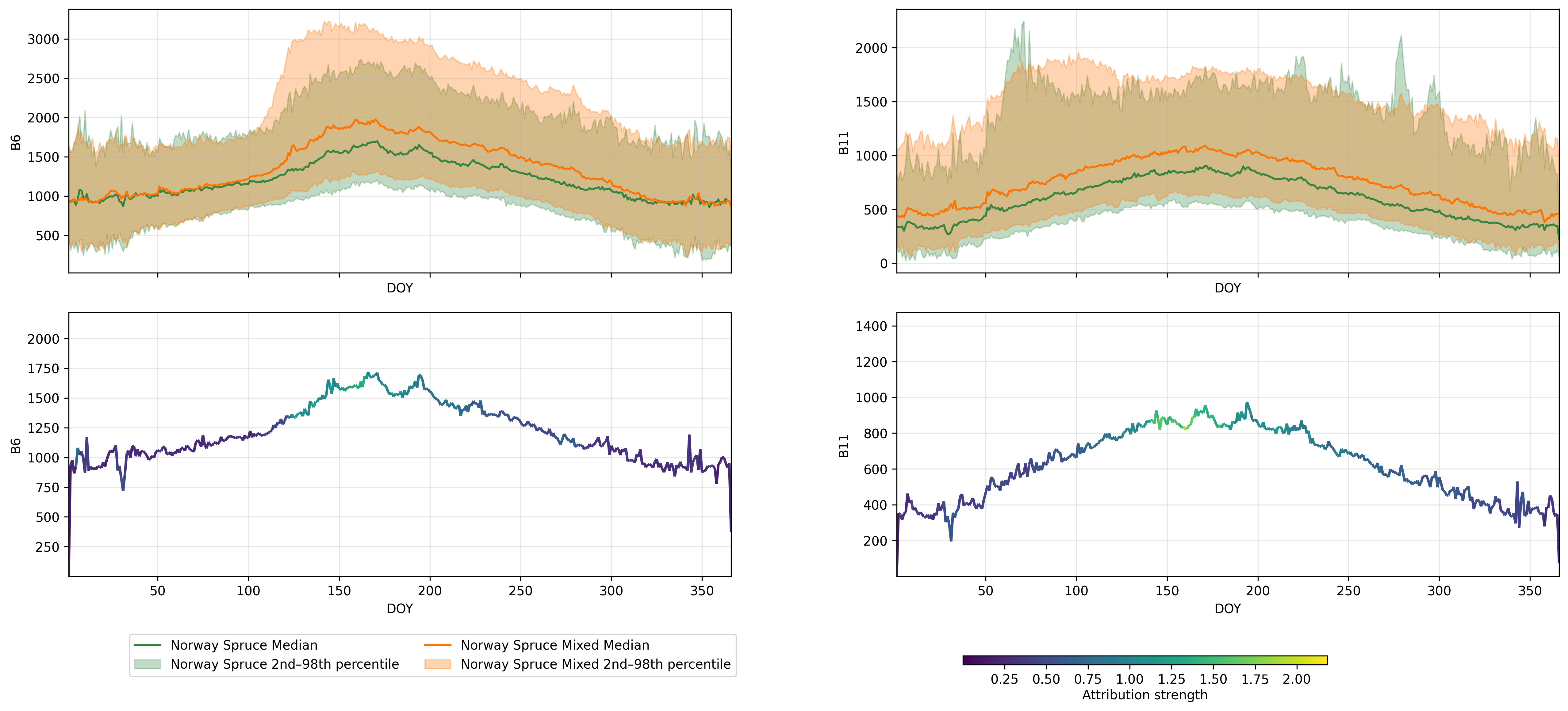}
    \caption{\textit{Separation between pure and mixed Norway spruce.} The top row shows the median test-set time series with confidence bands. The bottom row displays Integrated Gradients attributions along the Norway spruce median time series, highlighting time steps relevant for distinguishing the two species.}
    \label{fig:nwsvsmwsMixed_traj}
\end{figure}

The class-comparison \gls{ig} analysis provided insights beyond the zero-baseline analysis. Relative to a zero baseline, B6 showed the highest attributions for both pure and mixed Norway spruce. However, when each class was evaluated against a baseline derived from the other class, B11 became the most strongly attributed band. Thus, the features supporting individual class predictions differed from those used to distinguish the two closely related classes.
The direct comparison also revealed a seasonal pattern that was not apparent in the zero-baseline results. Attributions for B6, B7, B8 and B8A increased after approximately \gls{doy} 125, broadly coinciding with leaf development in Central European broadleaf species \citep{schieber_phenology_2009}. The differences become clearer when the median reflectance time series of pure and mixed Norway spruce are compared (see Figure~\ref{fig:nwsvsmwsMixed_traj}). The median B6 reflectance trajectories of the two classes begin to diverge around \gls{doy} 100 and remain separated for most of the growing season. However, the highest attribution values during this period occur in B11 rather than B6. In particular, B11 attributions increased when the difference between the B6 trajectories was most pronounced. B11 reflectance also remained consistently higher for the mixed class throughout the year, with a peak in late spring. These results suggest that the model did not rely on individual bands independently but rather on their combined spectral-temporal patterns when distinguishing pure from mixed Norway spruce stands. 

This result is particularly relevant because pure and mixed Norway spruce showed the highest degree of mutual misclassification across the evaluated models. Although the generally high relevance of red-edge and short-wave infrared bands is consistent with previous remote-sensing research \citep{wegler_tree_2025, immitzer_optimal_2019}, the class-comparison temporal attribution patterns provide more detailed insight into how the models distinguished these closely related forest classes.

Although the phenological patterns of beech and oak appeared more similar than those of pure and mixed Norway spruce, the models distinguished the two broadleaf species with high accuracy. However, the attribution patterns did not correspond directly to the periods with the largest visible differences between the median time series. Instead, the highest attributions occurred mainly between \gls{doy} 100 and 150, coinciding with the leaf-unfolding period. Although the median time series showed little separation between beech and oak during this period, the model appears to have used subtle spectral–temporal differences to distinguish the two species (Figure~\ref{fig:beech_v_oak}).

\begin{figure}[H]
    \centering
    \begin{adjustbox}{max width=1\columnwidth}
        \input{figs/oak_beech_b3_b6_b11_MHA}
    \end{adjustbox}
       \caption{\textit{Seperation between Beech and Oak}. The top row shows the median test-set time series with confidence bands. The bottom row displays Integrated Gradients attributions along the beech median time series, highlighting time steps relevant for distinguishing the two species.}
    \label{fig:beech_v_oak}
\end{figure}

\subsection{Potential Applications in Forestry}

\begin{figure}[H]
    \centering
     \begin{adjustbox}{max width=1\columnwidth}
            \input{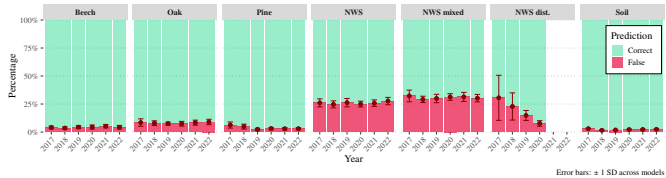}
            \end{adjustbox}
            \caption{\textit{False Predictions over years:} Here the mean percentage $\pm$ SD of false predictions for each class is visualized for every observed year across all RNN models. Norway spruce is abbreviated to \textit{NWS}. The disturbed Norway spruce class has no observations for the years 2021 and 2022, as there was no data where the disturbance year was after 2020.
                }
    \label{fig:fp_over_years_all}
\end{figure}

We expected the spectral-temporal patterns of species to vary severely between years, potentially reducing the effectiveness of single-year time series. In particular, extreme summers such as 2018 and 2022 in Germany were expected to alter spectral patterns and increase misclassification. However, the proportion of false predictions did not increase noticeably in these years (Fig.~\ref{fig:fp_over_years_all}). 

\begin{figure}[H]
    \centering
     \begin{adjustbox}{max width=1\columnwidth}
            \input{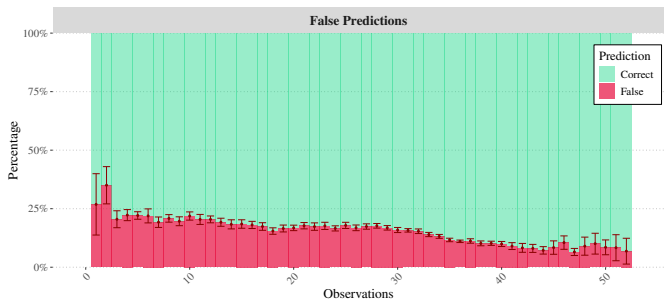}
            \end{adjustbox}
        \caption{\textit{Mean percentage of false predictions ($\pm$ SD) for pure and mixed Norway spruce combined, grouped by the number of time steps per time series.}}
    \label{fig:fp_over_obs}
\end{figure}

Across all RNN models, fewer annual observations generally reduced classification performance. The impact of observation count on correct classification percentage for pure and mixed Norway spruce over all RNN models is visualized in Figure ~\ref{fig:fp_over_obs}. This offers potential for future studues, as the launch of Sentinel-2C increased yearly observations in europe by about 33\% and reduced the average datagap lenght \citep{lewinska_benefits_2026}, which suggests better interpretability with our approach.     

A direct comparison with studies using multi-year time series for similar forest classes in Germany is difficult, as these studies often included a larger number of classes  \citep{wegler_tree_2025,blickensdorfer_national_2024,freudenberg_consistent_2025}. This increases the potential for misclassification, particularly among broadleaf species such as beech and oak. At the same time, other studies did not distinguish between pure and mixed Norway spruce stands, which likely contributed to higher Macro-F1 scores for Norway spruce. The classification tasks therefore differ substantially, and a robust comparison would require further targeted studies.
Nevertheless, the strong classification performance achieved here indicates that the appropriate input length is task-specific and that single-year time series should not be excluded by default. For detailed forest mapping, longer multi-year time series may provide higher classification performance \citep{wegler_tree_2025,blickensdorfer_national_2024,freudenberg_consistent_2025}. In contrast, shorter time series may be preferable when lightweight operational models or energy-efficient training and inference are priorities. Previous studies have shown that reducing the number of input observations can improve computational and energy efficiency \citep{verdecchia_data-centric_2022}.
In such a setting, reducing the number of input features also becomes important, as it further contributes to better energy efficiency \citep{verdecchia_data-centric_2022}. The \gls{ig} analysis indicated that bands B2 and B12 contributed little to the model predictions and could therefore be tested for removal to improve efficiency. Reducing the input dimensionality may also allow similar performance with smaller hidden representations and fewer trainable parameters, thereby lowering computational demand and associated emissions. However, this would require further evaluation.
Identifying the temporal and spectral observations that contribute most strongly to distinguishing pure and mixed stands provides useful information for forest composition mapping. The insights of the explainability analysis could guide the development of future models that better separate pure and mixed stands by focusing on the most informative spectral bands and seasonal periods. This may also enable feature and time-step reduction, supporting the development of more lightweight and computationally efficient models.

\subsection{Limitations}
The training dataset included observations from different locations and years, allowing the models to capture spatial and interannual variability. However, this variability was limited to the period from 2017 to 2022. In addition, the dataset providers noted that the dataset is well suited for model training but less suitable for independent validation \citep{freudenberg_sentinel-2_2025}. The reported classification performance should therefore be interpreted with caution. Furhtermore, Spatial and temporal autocorrelation may have influenced the performance estimates despite measures taken to prevent data leakage. To avoid any data leakage a leave-one-region-out cross-validation approach would be most appropriate. In this setup the data is spatially divided into regions and the model is trained on all but one region. The left out region is used for testing. This is repeated iteratively until each region has served once as the test set. 
We partly addressed these limitations by applying the model to an independent mixed-forest region. However, this provided only a limited visual assessment of model performance and cannot replace evaluation on a larger and more diverse independent test dataset.   
Another limitation is that the study region is limited to Germany and the results may not transfer directly to other forest types, climatic conditions or management systems.  We selected species classes based on the four predominant tree species in Germany, with different health stages and structures for Norway spruce, which also sets us apart from other german studies in terms of our class definitions and study design. This makes it difficult to draw direct comparisons with previous studies regarding model performance, particularly with regard to the separability of pure and mixed spruce stands, as other studies have not focused on this approach.
Finally, the potential efficiency gains from shorter time series, feature reduction, and smaller model architectures were not measured directly. Their effects on computational demand, energy consumption, and classification performance therefore require further evaluation.

\section{Conclusion}

This study investigated the classification and interpretation of annual Sentinel-2 time series for forest species mapping, with particular emphasis on the discrimination of spectrally and phenologically similar classes. Recurrent deep learning models were compared with established tree-based ensemble methods, while their spectral-temporal decision processes were examined using Integrated Gradients. The best-performing model, GRU-MHA-128, achieved a Macro-F1 score of 0.851, corresponding to only a modest improvement of 0.012 over XGBoost. The primary aim of this study, however, was not to optimize an operational tree species mapping approach, but to investigate which spectral bands and temporal periods contribute to model predictions and class separation.

Combining recurrent models with Integrated Gradients enabled attribution for each individual spectral band for each available timestep without prior temporal aggregation. In addition to conventional class-wise attributions, using observations from an opposing class as the baseline allowed us to identify the features driving class separation, rather than only those contributing to the prediction of an individual class. For pure and mixed Norway spruce, the contributions increased for B6, B7, B8, and B8A after approximately DOY 125, while B11 showed particularly strong attribution around DOY 100. For beech and oak, B6 and B11 and the period between approximately DOY 110 and 150 were especially relevant for class separation. The analysis further demonstrated that features contributing strongly to the prediction of an individual class are not necessarily identical to those driving the distinction between similar classes.

The spectral temporal attribution patterns identified for pure and mixed Norway spruce broadly corresponded with known phenological processes, particularly the development of broadleaf vegetation within mixed stands. The explainability analysis therefore provides a physically plausible interpretation of the model process, although it did not reveal substantially different spectral-temporal relationships beyond those expected from existing phenological knowledge. Still, explainable AI implications like \gls{ig} improve the understanding of the discrimination process and can inform future features or data design.

The increased number of annual observations offered by Sentinel-2C motivates repeating the proposed workflow with denser time series in future work, as more observations improved classification performance in this study. Additional temporal information may also reveal spectral-temporal relationships that are currently difficult to resolve. Furthermore, moving beyond categorical pure-versus-mixed labels toward estimating species proportions or subpixel forest composition could provide a more continuous representation of mixed stands and allow a more detailed investigation of the spectral response to increasing species admixture. 
Overall, this study demonstrates that recurrent deep learning models combined with class-comparative attribution provide a framework for investigating spectral-temporal decision processes in forest classification, particularly where classes are spectrally and phenologically similar.


\bibliography{references}

@article{lewinska_benefits_2026,
	title = {Benefits of the {Sentinel}-2 mission triplet constellation – one year on},
	volume = {44},
	issn = {23529385},
	url = {https://linkinghub.elsevier.com/retrieve/pii/S2352938526004118},
	doi = {10.1016/j.rsase.2026.102278},
	language = {en},
	urldate = {2026-10-11},
	journal = {Remote Sensing Applications: Society and Environment},
	author = {Lewińska, Katarzyna Ewa and Hagolle, Olivier and Gascon, Ferran and Griffiths, Patrick and Leser, Ulf and Hostert, Patrick},
	month = dec,
	year = {2026},
	pages = {102278},
}

@misc{thunen-institut_bundeswaldinventur_2024,
	title = {Bundeswaldinventur 2022 – {Ergebnisdatenbank}: {Vorrat} nach {Land} und {Baumartengruppe}},
	url = {https://bwi.info/inhalt1.3.aspx?Text=3.03%20Baumartengruppe&prRolle=public&prInv=BWI2022&prKapitel=3.03},
	urldate = {2026-08-20},
	author = {{Thünen-Institut}},
	year = {2024},
}

@incollection{seneviratne_weather_2021,
	address = {Cambridge, United Kingdom and New York, NY, USA},
	title = {Weather and {Climate} {Extreme} {Events} in a {Changing} {Climate}},
	doi = {10.1017/9781009157896.013},
	booktitle = {Climate {Change} 2021: {The} {Physical} {Science} {Basis}. {Contribution} of {Working} {Group} {I} to the {Sixth} {Assessment} {Report} of the {Intergovernmental} {Panel} on {Climate} {Change}},
	publisher = {Cambridge University Press},
	author = {Seneviratne, S. I. and Zhang, X. and Adnan, M. and Badi, W. and Dereczynski, C. and Di Luca, A. and Ghosh, S. and Iskandar, I. and Kossin, J. and Lewis, S. and Otto, F. and Pinto, I. and Satoh, M. and Vicente-Serrano, S. M. and Wehner, M. and Zhou, B.},
	editor = {Masson-Delmotte, V. and Zhai, P. and Pirani, A. and Connors, S. L. and Péan, C. and Berger, S. and Caud, N. and Chen, Y. and Goldfarb, L. and Gomis, M. I. and Huang, M. and Leitzell, K. and Lonnoy, E. and Matthews, J. B. R. and Maycock, T. K. and Waterfield, T. and Yelekçi, O. and Yu, R. and Zhou, B.},
	year = {2021},
	pages = {1513--1766},
}

@misc{bundesministerium_fur_landwirtschaft_ernahrung_und_heimat_bundeswaldinventur_2022,
	title = {Bundeswaldinventur 2022},
	url = {https://www.thuenen.de/de/themenfelder/waelder/die-bundeswaldinventur/ergebnisse-der-bundeswaldinventur-2022},
	journal = {Ergebnisse der vierten Bundeswaldinventur},
	author = {Bundesministerium für Landwirtschaft, Ernährung und Heimat},
	year = {2022},
}

@article{yachi_biodiversity_1999,
	title = {Biodiversity and ecosystem productivity in a fluctuating environment: {The} insurance hypothesis},
	volume = {96},
	issn = {0027-8424, 1091-6490},
	shorttitle = {Biodiversity and ecosystem productivity in a fluctuating environment},
	url = {https://pnas.org/doi/full/10.1073/pnas.96.4.1463},
	doi = {10.1073/pnas.96.4.1463},
	language = {en},
	number = {4},
	urldate = {2026-08-17},
	journal = {Proceedings of the National Academy of Sciences},
	author = {Yachi, Shigeo and Loreau, Michel},
	month = feb,
	year = {1999},
	pages = {1463--1468},
}

@inproceedings{verdecchia_data-centric_2022,
	address = {Plovdiv, Bulgaria},
	title = {Data-{Centric} {Green} {AI} {An} {Exploratory} {Empirical} {Study}},
	copyright = {https://doi.org/10.15223/policy-029},
	isbn = {9781665482868},
	url = {https://ieeexplore.ieee.org/document/9830097/},
	doi = {10.1109/ICT4S55073.2022.00015},
	urldate = {2026-08-05},
	booktitle = {2022 {International} {Conference} on {ICT} for {Sustainability} ({ICT4S})},
	publisher = {IEEE},
	author = {Verdecchia, Roberto and Cruz, Luis and Sallou, June and Lin, Michelle and Wickenden, James and Hotellier, Estelle},
	month = jun,
	year = {2022},
	pages = {35--45},
}

@article{schieber_phenology_2009,
	title = {Phenology of four broad-leaved forest trees in a submountain beech forest},
	volume = {55},
	copyright = {https://creativecommons.org/licenses/by-nc/4.0/},
	issn = {12124834, 1805935X},
	url = {http://jfs.agriculturejournals.cz/doi/10.17221/51/2008-JFS.html},
	doi = {10.17221/51/2008-JFS},
	number = {1},
	urldate = {2026-08-04},
	journal = {Journal of Forest Science},
	author = {Schieber, B. and Janík, R. and Snopková, Z.},
	month = jan,
	year = {2009},
	pages = {15--22},
}

@inproceedings{cho_learning_2014,
	address = {Doha, Qatar},
	title = {Learning {Phrase} {Representations} using {RNN} {Encoder}–{Decoder} for {Statistical} {Machine} {Translation}},
	url = {http://aclweb.org/anthology/D14-1179},
	doi = {10.3115/v1/D14-1179},
	language = {en},
	urldate = {2026-08-02},
	booktitle = {Proceedings of the 2014 {Conference} on {Empirical} {Methods} in {Natural} {Language} {Processing} ({EMNLP})},
	publisher = {Association for Computational Linguistics},
	author = {Cho, Kyunghyun and Van Merrienboer, Bart and Gulcehre, Caglar and Bahdanau, Dzmitry and Bougares, Fethi and Schwenk, Holger and Bengio, Yoshua},
	year = {2014},
	pages = {1724--1734},
}

@misc{lee_set_2018,
	title = {Set {Transformer}: {A} {Framework} for {Attention}-based {Permutation}-{Invariant} {Neural} {Networks}},
	copyright = {arXiv.org perpetual, non-exclusive license},
	shorttitle = {Set {Transformer}},
	url = {https://arxiv.org/abs/1810.00825},
	doi = {10.48550/ARXIV.1810.00825},
	urldate = {2026-07-23},
	publisher = {arXiv},
	author = {Lee, Juho and Lee, Yoonho and Kim, Jungtaek and Kosiorek, Adam R. and Choi, Seungjin and Teh, Yee Whye},
	year = {2018},
}

@incollection{pretzsch_mixed-species_2017,
	address = {Berlin, Heidelberg},
	title = {Mixed-{Species} {Forests}: {The} {Development} of a {Forest} {Management} {Paradigm}},
	isbn = {9783662545515 9783662545539},
	shorttitle = {Mixed-{Species} {Forests}},
	url = {http://link.springer.com/10.1007/978-3-662-54553-9_1},
	language = {en},
	urldate = {2026-07-20},
	booktitle = {Mixed-{Species} {Forests}},
	publisher = {Springer Berlin Heidelberg},
	author = {Bauhus, Jürgen and Forrester, David I. and Pretzsch, Hans},
	editor = {Pretzsch, Hans and Forrester, David I. and Bauhus, Jürgen},
	year = {2017},
	doi = {10.1007/978-3-662-54553-9_1},
	pages = {1--25},
}

@article{martone_global_2018,
	title = {The global forest/non-forest map from {TanDEM}-{X} interferometric {SAR} data},
	volume = {205},
	issn = {00344257},
	url = {https://linkinghub.elsevier.com/retrieve/pii/S0034425717305795},
	doi = {10.1016/j.rse.2017.12.002},
	language = {en},
	urldate = {2026-07-20},
	journal = {Remote Sensing of Environment},
	author = {Martone, Michele and Rizzoli, Paola and Wecklich, Christopher and González, Carolina and Bueso-Bello, José-Luis and Valdo, Paolo and Schulze, Daniel and Zink, Manfred and Krieger, Gerhard and Moreira, Alberto},
	month = feb,
	year = {2018},
	pages = {352--373},
}

@article{iglseder_potential_2023,
	title = {The potential of combining satellite and airborne remote sensing data for habitat classification and monitoring in forest landscapes},
	volume = {117},
	issn = {15698432},
	url = {https://linkinghub.elsevier.com/retrieve/pii/S1569843222003193},
	doi = {10.1016/j.jag.2022.103131},
	language = {en},
	urldate = {2026-07-20},
	journal = {International Journal of Applied Earth Observation and Geoinformation},
	author = {Iglseder, Anna and Immitzer, Markus and Dostálová, Alena and Kasper, Andreas and Pfeifer, Norbert and Bauerhansl, Christoph and Schöttl, Stefan and Hollaus, Markus},
	month = mar,
	year = {2023},
	pages = {103131},
}

@article{stoffels_satellite-based_2012,
	title = {Satellite-based stand-wise forest cover type mapping using a spatially adaptive classification approach},
	volume = {131},
	issn = {1612-4677},
	url = {https://doi.org/10.1007/s10342-011-0577-2},
	doi = {10.1007/s10342-011-0577-2},
	language = {en},
	number = {4},
	urldate = {2026-07-20},
	journal = {European Journal of Forest Research},
	author = {Stoffels, Johannes and Mader, Sebastian and Hill, Joachim and Werner, Willy and Ontrup, Godehard},
	month = jul,
	year = {2012},
	pages = {1071--1089},
}

@misc{freudenberg_consistent_2025,
	title = {Consistent {Multi}-{Temporal} {National} {Scale} {Tree} {Species} {Maps} for {Germany} 2018 – 2024},
	url = {https://www.ssrn.com/abstract=5705254},
	doi = {10.2139/ssrn.5705254},
	urldate = {2026-07-20},
	author = {Freudenberg, Maximilian and Schnell, Sebastian and Henning, Marko and Magdon, Paul},
	year = {2025},
}

@article{phiri_sentinel-2_2020,
	title = {Sentinel-2 {Data} for {Land} {Cover}/{Use} {Mapping}: {A} {Review}},
	volume = {12},
	issn = {2072-4292},
	shorttitle = {Sentinel-2 {Data} for {Land} {Cover}/{Use} {Mapping}},
	url = {https://www.mdpi.com/2072-4292/12/14/2291},
	doi = {10.3390/rs12142291},
	language = {en},
	number = {14},
	urldate = {2026-06-24},
	journal = {Remote Sensing},
	author = {Phiri, Darius and Simwanda, Matamyo and Salekin, Serajis and Nyirenda, Vincent and Murayama, Yuji and Ranagalage, Manjula},
	month = jul,
	year = {2020},
	pages = {2291},
}

@article{mouret_tree_2025,
	title = {Tree {Species} {Classification} at the {Pixel} {Level} {Using} {Deep} {Learning} and {Multispectral} {Time} {Series} in an {Imbalanced} {Context}},
	volume = {17},
	issn = {2072-4292},
	url = {https://www.mdpi.com/2072-4292/17/7/1190},
	doi = {10.3390/rs17071190},
	language = {en},
	number = {7},
	urldate = {2026-06-12},
	journal = {Remote Sensing},
	author = {Mouret, Florian and Morin, David and Planells, Milena and Vincent-Barbaroux, Cécile},
	month = mar,
	year = {2025},
	pages = {1190},
}

@article{blickensdorfer_national_2024,
	title = {National tree species mapping using {Sentinel}-1/2 time series and {German} {National} {Forest} {Inventory} data},
	volume = {304},
	issn = {00344257},
	url = {https://linkinghub.elsevier.com/retrieve/pii/S0034425724000804},
	doi = {10.1016/j.rse.2024.114069},
	language = {en},
	urldate = {2026-06-12},
	journal = {Remote Sensing of Environment},
	author = {Blickensdörfer, Lukas and Oehmichen, Katja and Pflugmacher, Dirk and Kleinschmit, Birgit and Hostert, Patrick},
	month = apr,
	year = {2024},
	pages = {114069},
}

@article{immitzer_optimal_2019,
	title = {Optimal {Input} {Features} for {Tree} {Species} {Classification} in {Central} {Europe} {Based} on {Multi}-{Temporal} {Sentinel}-2 {Data}},
	volume = {11},
	issn = {2072-4292},
	url = {https://www.mdpi.com/2072-4292/11/22/2599},
	doi = {10.3390/rs11222599},
	language = {en},
	number = {22},
	urldate = {2026-06-12},
	journal = {Remote Sensing},
	author = {Immitzer, Markus and Neuwirth, Martin and Böck, Sebastian and Brenner, Harald and Vuolo, Francesco and Atzberger, Clement},
	month = nov,
	year = {2019},
	pages = {2599},
}

@article{immitzer_first_2016,
	title = {First {Experience} with {Sentinel}-2 {Data} for {Crop} and {Tree} {Species} {Classifications} in {Central} {Europe}},
	volume = {8},
	issn = {2072-4292},
	url = {https://www.mdpi.com/2072-4292/8/3/166},
	doi = {10.3390/rs8030166},
	language = {en},
	number = {3},
	urldate = {2026-06-12},
	journal = {Remote Sensing},
	author = {Immitzer, Markus and Vuolo, Francesco and Atzberger, Clement},
	month = feb,
	year = {2016},
	pages = {166},
}

@article{grabska_forest_2019,
	title = {Forest {Stand} {Species} {Mapping} {Using} the {Sentinel}-2 {Time} {Series}},
	volume = {11},
	issn = {2072-4292},
	url = {https://www.mdpi.com/2072-4292/11/10/1197},
	doi = {10.3390/rs11101197},
	language = {en},
	number = {10},
	urldate = {2026-06-12},
	journal = {Remote Sensing},
	author = {Grabska, Ewa and Hostert, Patrick and Pflugmacher, Dirk and Ostapowicz, Katarzyna},
	month = may,
	year = {2019},
	pages = {1197},
}

@article{zhong_review_2024,
	title = {A {Review}: {Tree} {Species} {Classification} {Based} on {Remote} {Sensing} {Data} and {Classic} {Deep} {Learning}-{Based} {Methods}},
	volume = {15},
	issn = {1999-4907},
	shorttitle = {A {Review}},
	url = {https://www.mdpi.com/1999-4907/15/5/852},
	doi = {10.3390/f15050852},
	language = {en},
	number = {5},
	urldate = {2026-06-11},
	journal = {Forests},
	author = {Zhong, Lihui and Dai, Zhengquan and Fang, Panfei and Cao, Yong and Wang, Leiguang},
	month = may,
	year = {2024},
	pages = {852},
}

@article{tan_leveraging_2025,
	title = {Leveraging {Sentinel}-1/2 time series and deep learning for accurate forest tree species mapping},
	volume = {8},
	issn = {2624-893X},
	url = {https://www.frontiersin.org/articles/10.3389/ffgc.2025.1599510/full},
	doi = {10.3389/ffgc.2025.1599510},
	urldate = {2026-06-11},
	journal = {Frontiers in Forests and Global Change},
	author = {Tan, Jun and Li, Jing and Ma, Tianyue and Yan, Xingguang and Huo, Ziye},
	month = jul,
	year = {2025},
	pages = {1599510},
}

@misc{vaswani_attention_2017,
	title = {Attention {Is} {All} {You} {Need}},
	copyright = {arXiv.org perpetual, non-exclusive license},
	url = {https://arxiv.org/abs/1706.03762},
	doi = {10.48550/ARXIV.1706.03762},
	urldate = {2026-06-10},
	publisher = {arXiv},
	author = {Vaswani, Ashish and Shazeer, Noam and Parmar, Niki and Uszkoreit, Jakob and Jones, Llion and Gomez, Aidan N. and Kaiser, Lukasz and Polosukhin, Illia},
	year = {2017},
}

@incollection{ancona_gradient-based_2019,
	address = {Cham},
	title = {Gradient-{Based} {Attribution} {Methods}},
	isbn = {978-3-030-28954-6},
	url = {https://doi.org/10.1007/978-3-030-28954-6_9},
	booktitle = {Explainable {AI}: {Interpreting}, {Explaining} and {Visualizing} {Deep} {Learning}},
	publisher = {Springer International Publishing},
	author = {Ancona, Marco and Ceolini, Enea and Öztireli, Cengiz and Gross, Markus},
	editor = {Samek, Wojciech and Montavon, Grégoire and Vedaldi, Andrea and Hansen, Lars Kai and Müller, Klaus-Robert},
	year = {2019},
	doi = {10.1007/978-3-030-28954-6_9},
	pages = {169--191},
}

@inproceedings{sanyal_discretized_2021,
	address = {Online and Punta Cana, Dominican Republic},
	title = {Discretized {Integrated} {Gradients} for {Explaining} {Language} {Models}},
	url = {https://aclanthology.org/2021.emnlp-main.805},
	doi = {10.18653/v1/2021.emnlp-main.805},
	language = {en},
	urldate = {2026-05-29},
	booktitle = {Proceedings of the 2021 {Conference} on {Empirical} {Methods} in {Natural} {Language} {Processing}},
	publisher = {Association for Computational Linguistics},
	author = {Sanyal, Soumya and Ren, Xiang},
	year = {2021},
	pages = {10285--10299},
}

@article{jha_enhanced_2020,
	title = {Enhanced {Integrated} {Gradients}: improving interpretability of deep learning models using splicing codes as a case study},
	volume = {21},
	issn = {1474-760X},
	shorttitle = {Enhanced {Integrated} {Gradients}},
	url = {https://genomebiology.biomedcentral.com/articles/10.1186/s13059-020-02055-7},
	doi = {10.1186/s13059-020-02055-7},
	language = {en},
	number = {1},
	urldate = {2026-05-29},
	journal = {Genome Biology},
	author = {Jha, Anupama and K. Aicher, Joseph and R. Gazzara, Matthew and Singh, Deependra and Barash, Yoseph},
	month = dec,
	year = {2020},
	pages = {149},
}

@article{tronchin_translating_2024,
	title = {Translating {Image} {XAI} to {Multivariate} {Time} {Series}},
	volume = {12},
	copyright = {https://creativecommons.org/licenses/by-nc-nd/4.0/},
	issn = {2169-3536},
	url = {https://ieeexplore.ieee.org/document/10439172/},
	doi = {10.1109/ACCESS.2024.3366994},
	urldate = {2026-05-29},
	journal = {IEEE Access},
	author = {Tronchin, Lorenzo and Cordelli, Ermanno and Celsi, Lorenzo Ricciardi and Maccagnola, Daniele and Natale, Massimo and Soda, Paolo and Sicilia, Rosa},
	year = {2024},
	pages = {27484--27500},
}

@misc{samek_explainable_2017,
	title = {Explainable {Artificial} {Intelligence}: {Understanding}, {Visualizing} and {Interpreting} {Deep} {Learning} {Models}},
	copyright = {arXiv.org perpetual, non-exclusive license},
	shorttitle = {Explainable {Artificial} {Intelligence}},
	url = {https://arxiv.org/abs/1708.08296},
	doi = {10.48550/ARXIV.1708.08296},
	urldate = {2026-05-29},
	publisher = {arXiv},
	author = {Samek, Wojciech and Wiegand, Thomas and Müller, Klaus-Robert},
	year = {2017},
}

@article{hassija_interpreting_2024,
	title = {Interpreting {Black}-{Box} {Models}: {A} {Review} on {Explainable} {Artificial} {Intelligence}},
	volume = {16},
	issn = {1866-9956, 1866-9964},
	shorttitle = {Interpreting {Black}-{Box} {Models}},
	url = {https://link.springer.com/10.1007/s12559-023-10179-8},
	doi = {10.1007/s12559-023-10179-8},
	language = {en},
	number = {1},
	urldate = {2026-05-29},
	journal = {Cognitive Computation},
	author = {Hassija, Vikas and Chamola, Vinay and Mahapatra, Atmesh and Singal, Abhinandan and Goel, Divyansh and Huang, Kaizhu and Scardapane, Simone and Spinelli, Indro and Mahmud, Mufti and Hussain, Amir},
	month = jan,
	year = {2024},
	pages = {45--74},
}

@inproceedings{malkanthi_forest_2025,
	address = {Colombo, Sri Lanka},
	title = {Forest {Cover} {Classification} with {Explainable} {AI}: {A} {Comparative} {Analysis}},
	copyright = {https://doi.org/10.15223/policy-029},
	isbn = {9798331561239},
	shorttitle = {Forest {Cover} {Classification} with {Explainable} {AI}},
	url = {https://ieeexplore.ieee.org/document/11318453/},
	doi = {10.1109/SLAAI-ICAI68534.2025.11318453},
	urldate = {2026-05-22},
	booktitle = {2025 9th {SLAAI} {International} {Conference} on {Artificial} {Intelligence} ({SLAAI}-{ICAI})},
	publisher = {IEEE},
	author = {Malkanthi, A.M.C. and Rathnayaka, R.M.K.T. and Kumara, B.T.G.S. and Welikanna, D.R.},
	month = nov,
	year = {2025},
	pages = {1--6},
}

@article{liu_evaluating_2024,
	title = {Evaluating the {Potential} of {Sentinel}-2 {Time} {Series} {Imagery} and {Machine} {Learning} for {Tree} {Species} {Classification} in a {Mountainous} {Forest}},
	volume = {16},
	issn = {2072-4292},
	url = {https://www.mdpi.com/2072-4292/16/2/293},
	doi = {10.3390/rs16020293},
	language = {en},
	number = {2},
	urldate = {2026-05-15},
	journal = {Remote Sensing},
	author = {Liu, Pan and Ren, Chunying and Wang, Zongming and Jia, Mingming and Yu, Wensen and Ren, Huixin and Xia, Chenzhen},
	month = jan,
	year = {2024},
	pages = {293},
}

@article{abdi_mapping_2026,
	title = {Mapping forest tree species and their uncertainty using {Earth} observation and {National} {Forest} {Inventory} data: towards operational monitoring in {Sweden}},
	volume = {47},
	issn = {0143-1161, 1366-5901},
	shorttitle = {Mapping forest tree species and their uncertainty using {Earth} observation and {National} {Forest} {Inventory} data},
	url = {https://www.tandfonline.com/doi/full/10.1080/01431161.2026.2625513},
	doi = {10.1080/01431161.2026.2625513},
	language = {en},
	number = {7},
	urldate = {2026-05-15},
	journal = {International Journal of Remote Sensing},
	author = {Abdi, Abdulhakim M. and Wang, Fan},
	month = apr,
	year = {2026},
	pages = {2912--2943},
}

@article{wegler_tree_2025,
	title = {Tree species from space: a new product for {Germany} based on {Sentinel}-1 and -2 time series},
	volume = {46},
	issn = {0143-1161, 1366-5901},
	shorttitle = {Tree species from space},
	url = {https://www.tandfonline.com/doi/full/10.1080/01431161.2025.2530236},
	doi = {10.1080/01431161.2025.2530236},
	language = {en},
	number = {16},
	urldate = {2026-05-15},
	journal = {International Journal of Remote Sensing},
	author = {Wegler, Marco and Kacic, Patrick and Thonfeld, Frank and Holzwarth, Stefanie and Jaggy, Niklas and Gessner, Ursula and Kuenzer, Claudia},
	month = aug,
	year = {2025},
	pages = {6075--6108},
}

@article{kirsch_early_2025,
	title = {Early {Detection} of {Forest} {Disturbances} in {Homogeneous} {Stands} – {Deep} {Learning}-{Based} {Monitoring} of \textit{{Picea} {Abies}}},
	volume = {18},
	copyright = {https://creativecommons.org/licenses/by/4.0/legalcode},
	issn = {1939-1404, 2151-1535},
	url = {https://ieeexplore.ieee.org/document/11141690/},
	doi = {10.1109/JSTARS.2025.3603094},
	urldate = {2026-02-02},
	journal = {IEEE Journal of Selected Topics in Applied Earth Observations and Remote Sensing},
	author = {Kirsch, Maximilian and Wernicke, Jakob and Datta, Pawanjeet Singh and Preisach, Christine},
	year = {2025},
	pages = {22300--22316},
}

@book{russell_artificial_2020,
	address = {Hoboken, NJ},
	edition = {Fourth Edition},
	series = {Pearson {Series} in {Artificial} {Intelligence}},
	title = {Artificial intelligence: a modern approach},
	isbn = {978-0-13-461099-3},
	shorttitle = {Artificial intelligence},
	publisher = {Pearson},
	author = {Russell, Stuart J. and Norvig, Peter},
	year = {2020},
}

@article{freudenberg_sentinel-2_2025,
	title = {A {Sentinel}-2 machine learning dataset for tree species classification in {Germany}},
	volume = {17},
	copyright = {https://creativecommons.org/licenses/by/4.0/},
	issn = {1866-3516},
	url = {https://essd.copernicus.org/articles/17/351/2025/},
	doi = {10.5194/essd-17-351-2025},
	language = {en},
	number = {2},
	urldate = {2025-09-23},
	journal = {Earth System Science Data},
	author = {Freudenberg, Maximilian and Schnell, Sebastian and Magdon, Paul},
	month = feb,
	year = {2025},
	pages = {351--367},
}

@misc{sundararajan_axiomatic_2017,
	title = {Axiomatic {Attribution} for {Deep} {Networks}},
	url = {https://arxiv.org/abs/1703.01365},
	doi = {10.48550/ARXIV.1703.01365},
	urldate = {2025-07-22},
	publisher = {arXiv},
	author = {Sundararajan, Mukund and Taly, Ankur and Yan, Qiqi},
	year = {2017},
}

@article{kong_long_2018,
	title = {Long {Short}-{Term} {Memory} {Neural} {Networks} for {Online} {Disturbance} {Detection} in {Satellite} {Image} {Time} {Series}},
	volume = {10},
	issn = {2072-4292},
	doi = {10.3390/rs10030452},
	number = {3},
	journal = {Remote Sensing},
	author = {Kong, Yun-Long and Huang, Qingqing and Wang, Chengyi and Chen, Jingbo and Chen, Jiansheng and He, Dongxu},
	month = mar,
	year = {2018},
	pages = {452},
}

@article{abdullah_timing_2019,
	title = {Timing of red-edge and shortwave infrared reflectance critical for early stress detection induced by bark beetle ({Ips} typographus, {L}.) attack},
	volume = {82},
	issn = {15698432},
	doi = {10.1016/j.jag.2019.101900},
	journal = {International Journal of Applied Earth Observation and Geoinformation},
	author = {Abdullah, Haidi and Skidmore, Andrew K. and Darvishzadeh, Roshanak and Heurich, Marco},
	month = oct,
	year = {2019},
	pages = {101900},
}

@article{hochreiter_long_1997,
	title = {Long {Short}-{Term} {Memory}},
	volume = {9},
	issn = {0899-7667, 1530-888X},
	doi = {10.1162/neco.1997.9.8.1735},
	number = {8},
	journal = {Neural Computation},
	author = {Hochreiter, Sepp and Schmidhuber, Jürgen},
	month = nov,
	year = {1997},
	pages = {1735--1780},
}

@book{european_commission_joint_research_centre_climate_2020,
	address = {LU},
	title = {Climate change impacts and adaptation in {Europe}: {JRC} {PESETA} {IV} final report.},
	shorttitle = {Climate change impacts and adaptation in {Europe}},
	url = {https://data.europa.eu/doi/10.2760/171121},
	language = {eng},
	urldate = {2024-08-25},
	publisher = {Publications Office},
	author = {{European Commission. Joint Research Centre.}},
	year = {2020},
}

\clearpage
\appendix
\section{Appendix}
\subsection{Equations for LSTM and GRU}
\label{app:lstmandgru_equations}

\begin{equation}
    f_t = \sigma\left(w_f x_t + u_f h_{t-1} + b_f\right) 
    \label{eq:forget_gate}
\end{equation}
\begin{equation}
    i_t = \sigma\left(w_i x_t + u_i h_{t-1} + b_i\right) 
    \label{eq:input_gate}
\end{equation}
\begin{equation}
    o_t = \sigma\left(w_o x_t + u_o h_{t-1} + b_o\right) 
    \label{eq:output_gate}
\end{equation}
\begin{equation}
    \tilde{C}_t = tanh \left(w_{\tilde{C}} x_t + u_{\tilde{C}} h_{t-1} + b_{\tilde{C}}\right) 
    \label{eq:ig_2}
\end{equation}
\begin{equation}
    C_t = f_t \odot C_{t-1}+i_t \odot \tilde{C}_t 
    \label{eq:updated_LTM}
\end{equation}
\begin{equation}
    h_t = o_t\odot tanh\left(C_t\right) 
    \label{eq:ht_out}
\end{equation}

Where:
\par 
\begin{itemize}
    \item The \emph{forget gate} is denoted \(f_t\), the \emph{input gate} \(i_t\), the \emph{cell candidate gate} \(\tilde{C}_t\) and the \emph{output gate} \(o_t\).
    \item The input weights are represented as \(w\), the recurrent weights for the Hidden State as \(u\) and the biases are denoted as \(b\). 
    \item \(h\) is the hidden state, \(C\) is the cell state and \(x\) is the input data at time step \(t\).
    \item \(\odot\) denotes the point-wise multiplication of two vectors.
    \item \(\sigma\) is the sigmoid activation function, \emph{tanh} is the tanh activation function. 
\end{itemize}

For the GRU unit the gates are defined by the following functions \citep{cho_learning_2014}:

\begin{equation}
    r_t =
    \sigma\left(
        w_{r}x_t + b_{r}
        + w_{r}h_{t-1} + b_{r}
    \right)
    \label{eq:gru_reset_gate}
\end{equation}
\begin{equation}
    z_t =
    \sigma\left(
        w_{z}x_t + b_{z}
        + w_{z}h_{t-1} + b_{z}
    \right)
    \label{eq:gru_update_gate}
\end{equation}
\begin{equation}
    g_t =
    \tanh\left(
        w_{g}x_t + b_{g}
        + r_t \odot
        \left(
            w_{g}h_{t-1} + b_{g}
        \right)
    \right)
    \label{eq:gru_candidate_hidden_state}
\end{equation}
\begin{equation}
    h_t =
    \left(1-z_t\right)\odot g_t
    + z_t\odot h_{t-1}
    \label{eq:gru_hidden_state}
\end{equation}

Where:
\par
\begin{itemize}
    \item The \emph{reset gate} is denoted by \(r_t\), the \emph{update gate} by \(z_t\), and the \emph{candidate hidden state} by \(g_t\).
\end{itemize}

The sigmoid activation function is calculated as: 

\begin{equation}
    \sigma(x) = \frac{1}{1 + e^{-x}}
    \label{eq:sigmoid}
\end{equation}
\par
The tanh activation function is calculated as:
\begin{equation}
    tanh(x) =  \frac{e^x - e^{-x}}{e^x + e^{-x}}
    \label{eq:tanh}
\end{equation}

\subsection{Additional Tables}

\begin{table*}[t]
  \centering
    \caption{Comparison of model configurations}
    \label{tab:model_comparison2}
        \renewcommand{\arraystretch}{0.95}
    \begin{tabular*}{0.7\textwidth}{@{\extracolsep{\fill}}lcccccc@{}}
        \toprule
        
        \multirow{2}{*}{Model}
        & \multirow{2}{*}{\textbf{RF}}
        & \multirow{2}{*}{\textbf{\gls*{xgb}}}
        & \multicolumn{2}{c}{\textbf{LSTM}}
        & \multicolumn{2}{c}{\textbf{GRU}}
        \\
        
        \cmidrule(lr{0.4em}){4-5}
        \cmidrule(l{0.4em}){6-7}
        
        &
        &
        &
        MHA
        & No Att.
        & MHA
        & No Att.
        \\
        
        \midrule
        
        Accuracy
        & 0.792
        & 0.851
        & 0.848
        & 0.850
        & \textbf{0.854}
        & 0.853
        \\
        
        Balanced accuracy
        & 0.890
        & 0.925
        & 0.930
        & 0.931
        & \textbf{0.933}
        & \textbf{0.933}
        \\
        
        Macro F1
        & 0.748
        & 0.839
        & 0.843
        & 0.843
        & \textbf{0.851}
        & 0.847
        \\
        
        \bottomrule
    \end{tabular*}
\end{table*}

\begin{table}[H]
\centering
\caption{Hyperparameter search spaces for Random Forest and \gls*{xgb} models}
\label{tab:hyperparameter}
      \small
    \begin{tabular}{lll}
    \toprule
\textbf{Model} & \textbf{Hyperparameter} & \textbf{Tested values} \\
\midrule
\multirow{8}{*}{Random Forest}
& \texttt{n\_estimators} & 100, 250, 500, 750 \\
& \texttt{max\_depth} & 3, 6, 10, 15 \\
& \texttt{min\_samples\_split} & 2, 5, 10 \\
& \texttt{min\_samples\_leaf} & 1, 2, 5 \\
& \texttt{max\_features} & \texttt{sqrt}, \texttt{log2}, 0.5, 0.7 \\
& \texttt{bootstrap} & \texttt{True} \\
& \texttt{max\_samples} & 0.5, 0.7, 0.9, \texttt{None} \\
& \texttt{class\_weight} & \texttt{None}, \texttt{balanced} \\
\midrule
\multirow{8}{*}{XGBoost}
& \texttt{max\_depth} & 3, 6, 10, 15 \\
& \texttt{learning\_rate} & 0.01, 0.1, 0.2, 0.3, 0.4 \\
& \texttt{subsample} & 0.5, 0.6, 0.7, 0.8, 0.9 \\
& \texttt{colsample\_bytree} & 0.5, 0.6, 0.7, 0.8, 0.9 \\
& \texttt{colsample\_bylevel} & 0.5, 0.6, 0.7, 0.8, 0.9 \\
& \texttt{n\_estimators} & 100, 250, 500, 750 \\
& \texttt{reg\_alpha} & 0.1, 0.001, 0.00001 \\
& \texttt{reg\_lambda} & 0.1, 0.001, 0.00001 \\
\midrule
\multirow{6}{*}{RNN-Models}
& \texttt{RNN} & LSTM, GRU \\
& \texttt{Attention} & MHA, None \\
& \texttt{Bi-directional} & True\\
& \texttt{Bottleneck dimension} & 32, 64, 128 \\
& \texttt{Hidden Dimension} & Bottleneck dimension$ \times 4$ \\
& \texttt{dropout} &  0.2 \\
\bottomrule
\end{tabular}
\end{table}

\end{document}